\documentclass{article}
\usepackage[colorlinks,urlcolor=blue,linkcolor=blue,citecolor=blue]{hyperref}
\usepackage{subfig}
\usepackage{gensymb}
\usepackage{color,array}
\usepackage{mathtools,amssymb,lipsum, nccmath}
\usepackage[noend]{algpseudocode}
\usepackage{algorithm}
\usepackage{algorithmicx}
\usepackage{graphicx}
\usepackage{amsmath}
\usepackage{soul}
\usepackage{multirow,tabularx}
\DeclareMathOperator*{\argmin}{arg\,min} 
\DeclareMathOperator*{\argmax}{arg\,max}

\usepackage{arxiv}
\usepackage[utf8]{inputenc} 
\usepackage[T1]{fontenc}    
\usepackage{hyperref}       
\usepackage{url}            
\usepackage{booktabs}       
\usepackage{amsfonts}       
\usepackage{nicefrac}       
\usepackage{microtype}      
\usepackage{lipsum}
\usepackage{graphicx}
\graphicspath{ {./images/} }
\usepackage{xcolor}
\usepackage{multirow}

\usepackage{textcomp}
\usepackage{gensymb}
\usepackage{amsmath}
\DeclareGraphicsExtensions{.pdf,.png}

\author{
 Shashi Suman$^1$, Francois Rivest$^2$, Ali Etemad$^1$ \\
  $^1$Dept. ECE and Ingenuity Labs Research Institute, Queen's University\\
  $^2$Dept. of Mathematics and Computer Science, Royal Military College of Canada\\
  \texttt{shashi.suman@queensu.ca,
  francois.rivest@\{mail.mcgill.ca, rmc.ca\},
  ali.etemad@queensu.ca}
  }

\begin{document}

\title{Towards Personalization of User Preferences in Partially Observable Smart Home Environments}

\maketitle

\begin{abstract}
The technologies used in smart homes have recently improved to learn the user preferences from feedback in order to enhance the user convenience and quality of experience. Most smart homes learn a uniform model to represent the thermal preferences of users, which generally fails when the pool of occupants includes people with different sensitivities to temperature, for instance due to age and physiological factors. Thus, a smart home with a single optimal policy may fail to provide comfort when a new user with a different preference is integrated into the home. In this paper, we propose a Bayesian Reinforcement learning framework that can approximate the current occupant state in a partially observable smart home environment using its thermal preference, and then identify the occupant as a new user or someone is already known to the system. Our proposed framework can be used to identify users based on the temperature and humidity preferences of the occupant when performing different activities to enable personalization and improve comfort. We then compare the proposed framework with a baseline long short-term memory learner that learns the thermal preference of the user from the sequence of actions which it takes. We perform these experiments with up to 5 simulated human models each based on hierarchical reinforcement learning. The results show that our framework can approximate the belief state of the current user just by its temperature and humidity preferences across different activities with a high degree of accuracy. 
\end{abstract}

\section{Introduction}


The adoption of Smart Homes Systems (SHS) has increased significantly in recent years. With the development of the Internet of Things, it has become much easier to control and automate various devices such as thermostats, speakers, voice assistants \cite{SmartThermostat, SmartThermostat2, SmartLight}, and many other devices with the overall aim of improving the experience of occupants in the home. 
As a result of advances in automation, machine learning, and biometrics, such devices are now able to learn user preference patterns, and thus improve security, safety, and adaptability and personalization toward the occupants. 

On the topic of improving the experience of occupants, SHS are also responsible for monitoring and controlling comfort factors such as temperature, humidity, and lighting in order to maximize occupant comfort. These parameters can vary among different people within a group based on their physical characteristics like gender, age, and clothing \cite{personalization6gender}. They can even vary for the same person at different times based on their activities, state of health \cite{covid}, and others.
This shows the importance of personalization within SHS to provide a tailored experience to the occupants, thereby reducing manual intervention. Having a personalized setting attracts consumers' interest and helps in collecting and analyzing data to improve the adaptability of the SHS to a higher degree. Another way to improve this is with a cross-exchange of suggestions and feedbacks \cite{teachRobots} between the SHS and the occupant. By following the directions of the user \cite{FollowSHS}, SHS can perceive the user state indirectly, thus helping the SHS to build a hierarchical decision tree that is exclusive to each user. To achieve this, however, the SHS need access to parameters that are exclusive to the occupant in order to learn the behavior of the user.

As argued above, the ambient parameters that are often controlled by smart home agents can vary for different users, and even for the same user when performing different activities. While smart home agents can generally learn the behaviour of individual users based on user identity information, activities, and preferences, such parameters may not always be observable to the smart home agent when a new user is introduced to the environment. In some cases, smart homes do not have the ability to identify occupants due to the absence of ubiquitous biometric technologies \cite{biometric1, biometric2} as well as privacy issues which can be encountered if sensing systems like cameras were used to identify and personalize occupant preferences.

Previously in \cite{Suman}, we simulated a series of experiments with a Hierarchical Reinforcement Learning based human model capable of learning activities and setting its thermal preferences in a smart home environment. The SHS in turn aimed to learn the preferences of the human to improve the thermal comfort of the occupant (human model). Nonetheless, in \cite{Suman}, the SHS was designed to have full observability about the user. This was done in order to enable the SHS to adapt to users with different intrinsic reward functions. In this paper, we tackle the problem of user personalization in a scenario where the user identity in not observable to the smart home. To compensate for the partial information available about the user, we build into our smart home agent the ability to learn to recognize the user through its  activity and thermal preferences in order to improve its policy. To accomplish the task of learning a policy with uncertainty, our agent maintains a belief over the human model who is pursuing its activity in the environment. Our experiments show that the SHS can learn to accurately identify the current user in the environment based on its thermal preferences given the current activity among a set of possible activities. Beyond identifying the occupant in the environment, the SHS also learns the optimal thermal preference for each activity for the identified user, which can be used for further personalization.

In summary, our contributions in this paper are as follows.
(\textbf{1}) We tackle the scenario of user personalization where user identity is not fully observable to the smart home system, and instead rely on limited ambient parameters such as temperature and humidity preferences.
(\textbf{2}) We introduce a Bayesian model of the smart home that can approximate the user state using the thermal preference of the user for a given activity.
(\textbf{3}) We test our model on scenarios where we have up to 5 human models pursuing different activities in a home. We demonstrate that our model can successfully recognize the user by means of their activity-specific thermal preference with a high accuracy while also optimizing their comfort.

The rest of this paper is organized as follows. In the following section we present a summary of the related work on user thermal personalization. Next, in Section 3, we describe the architecture of the smart home agent that can learn to recognize the current user and to set the temperature and humidity (TH) according to its preferences. In Section 4 we implement a long short-term memory learner as a baseline to which we compare the performance of our RL-based model.
Next, in Section 5, we simulate the environment with varying number of human occupants and evaluate the performance of our model in detail. 
Finally, in Section 6, we provide the concluding remarks and discuss the limitations of our work. We also provide a few insights on possible future directions.


\section{Related Work}
In this section, we review several important prior works that have explored personalization of user's preferences towards achieving thermal comfort. It has been previously shown that when multiple users with different thermal preferences are present in a smart home, the SHS needs to have a personalized profile for each occupant. For instance, in \cite{UserPersonalSkin1}, boosted trees were used to learn occupant's personalized thermal responses using skin temperature and its surrounding, achieving a median accuracy of 84\%. Similarly, in \cite{UserPersonal2}, occupants heating and cooling behaviors were used to design personalized comfort models, obtaining a prediction accuracy of 73\%. It was shown in \cite{PersonalizedHeatFlux} that human occupants have different thermal perceptions in indoor environments based on heat exchange through the skin. A Random Forest (RF) was used and a median accuracy of 70.8\% was achieved for classifying thermal preferences with humans in the loop using air temperature. In \cite{PersonalizedSensation}, a Support Vector Machine (SVM) model was used to learn and classify the individual occupant's thermal sensation indoors, obtaining an accuracy of up to 86\%.

Experiments in \cite{standard199255} showed that on average, the thermal sensation of humans have a standard deviation of 3$\degree$C. This can also be confirmed from \cite{Personalization3} where the thermal sensation of a fixed number of users showed that different users perceive the same surroundings in a different way, thus the difference in the thermal perception among the users can be used to personalize their preferences. The parameters used in \cite{Personalization3} were Electrodermal Activity, humidity, operational skin temperature, and heart rate, which were used to classify the users using an SVM. Similarly in \cite{Personalization4IOT}, the thermal prediction error of users was reduced by 50\% using an SVM.

While the above-mentioned works use sensors attached to the body to obtain personalized data from the users, factors such as age and gender can also play a role in thermal perception and thus personalization. In \cite{personalization5}, age and outside temperature were included in the model of \cite{standard199255} to obtain personalized comfort levels on a fixed number of occupants, obtaining a prediction accuracy of 76.7\% using an SVM. Similarly, in \cite{personalization6gender}, the parameter of gender  was included to estimate the comfort of individuals and concluded that difference in clothing and gender were mostly responsible for explaining difference in preferences. To implement a personalized preference model, authors in \cite{personalization8Random} used Random Forest models and obtained a mean accuracy of 75\% to predict personalized thermal parameters.

The mentioned research works converge on a common aim of suggesting that a `one-model-fits-all' strategy does not work well with all individuals. This notion was further explored in \cite{personalization7thermalvotes}, where it was found that the preferences of many users did not fit a basic thermal model of \cite{standard199255} using $k$-Nearest Neighbour. After including these users, the thermal comfort among the occupants improved significantly while the thermal model's \cite{standard199255} accuracy decreased. Similarly, in \cite{personalization8Random}, it was discovered that nearly 30\% of users showed discomfort when placed within the same thermal conditions. These studies imply that a personalized model is necessary to learn the optimal comfort of each user in a home. These studies have, however, focused more on a fixed number of individuals with specific sensors to measure information such as skin temperature, heart rate, cardiac rhythm, and others, which may not always be available to the SHS.

In contrast to the majority of the above-mentioned studies, in this paper we focus on proposing a system that can recognize the users by their preferences in a situation where the only available information are the temperature and humidity adjustments made by them.



\section{Methods}
In this section, we extend the RL-based SHS system from our previous work \cite{Suman} with the ability to estimate who is the current occupant using current TH and the human model's activity and actions only. The assumption for this problem is that there is only one human model in the environment at a given time.
In this section, we first describe the thermal comfort model and our approach to synthesizing data with simulations, followed by the general framework for the problem, and finally the description of our proposed solution.

\subsection{Human Thermal Comfort Model and Simulations}
In order to implement the human thermal comfort, we use the model presented in \cite{standard199255} that uses thermal data from a large number of participants. The model approximates human thermal comfort with a mean vote. The range of the vote, also called Predicted Mean Vote (PMV), lies between -3 to 3, where -3 denotes very cold feeling and 3 denotes very hot feeling. Here, -0.5 to 0.5 is the most comfortable range. With a discomfortability rate of $5\%$, this data also accounts for the deviation in the thermal preference of occupants. The data accounts for the parameters that affect the thermal feelings of a user like temperature, humidity, clothing, activity, etc. Similarly, to simulate human decision making ability using RL, we implement the methodology in \cite{gebhardt} that uses Hierarchical Reinforcement Learning (HRL) to simulate how humans learn to switch between different activities by anticipating rewards from current and external activities. The complete model description can be found in \cite{Suman}.

\subsection{Partially Observable Markov Decision Process}

When there is uncertainty about the current state state of the environment (in this case, who the current occupant is, or what its preference are), a common framework for RL is the Partially Observable Markov Decision Process (POMDP) \cite{POMDP_Bayes, POMDP_Bayes_2, POMDP_LSTM, POMDP_LSTM_2}. A POMDP is defined by the tuple $<~\mathcal{S}, \mathcal{A}, \Omega, T, R, O, \gamma, b_0>$, where $\mathcal{S}$ is the set of discretized states, $\mathcal{A}$ is the set of actions, and $\Omega$ is the set of observations. Moreover, $T(s_t, a_t, s_{t+1}): [S \times A \times S] \rightarrow [0, 1]$ is the transition function that defines the probability of ending in state $s_{t+1}$ after taking action $a_t$ in state $s_t$,  $R$ is the reward function that specifies the immediate reward received after taking action $a$ in state $s_t$, and $O(a_t, o_{t+1}, s_{t+1}): [A \times \Omega \times S]\rightarrow [0, 1]$ is the observation function that defines the probability of observing $o_{t+1}$ after taking action $a_t$ and ending in state $s_{t+1}$. Lastly, $\gamma$ is the discount factor and $b_0$ is the initial belief state that defines a probability over the observable state of the environment.


To calculate the belief about the current state, POMDP uses the Bayes theorem, which is defined as:
\begin{equation}
    P(A|B) = \frac{P(B|A) P(A)}{P(B)},
\end{equation}
where $A$ and $B$ are the events, $P(A)$ is the prior, $P(B)$ is the evidence, $P(A|B)$ is the posterior probability of event $A$ given that event $B$ has occurred, $P(B|A)$ is the probability of event $B$ given that event $A$ has occurred (likelihood). Here, we set $A$ as a possible human model $\mathcal{H}_i$ in the home,
and $B$ as the next TH and activity $o_{t+1}$ observed by the smart home. Note that here the state is the concatenation of the observation $o_{t+1}$ and the current occupant $\mathcal{H}_i$ ($s_t = [o_t, \mathcal{H}_i]$), but the belief only needs to be over the occupants (since the TH and activity are fully observable). Accordingly, $P(B|A)$ is the probability of observing TH preferences given user $\mathcal{H}_i$ from the observation function $O$. $P(A|B)$ is the posterior probability of the human occupant $\mathcal{H}_i$ given the current activity and TH of the environment and previous belief. Finally $P(A)$ is the prior probability of human occupant $\mathcal{H}_i$ derived from the transition function $T$, while $P(B)$ is the marginal probability of the current observation received from the environment. Given our problem setup, the smart home agent takes an action $a_t$ at every time-step after which it receives an observation $o_{t+1}$ (consisting of the current activity of the occupant, temperature, and humidity) from the environment. With this observation, a POMDP agent would update its belief over each possible occupant by the following equation:
\begin{equation}
\begin{aligned}
    & \overbrace{b_{t+1}(\mathcal{H}_i)}^{P(A|B)} = \\
    & \frac{\overbrace{O(a_t,  o_{t+1}, [o_{t+1}, \mathcal{H}_i])}^{P(B|A)}\overbrace{\sum_{\mathcal{H}_j \in \mathcal{H}} T([o_t, \mathcal{H}_j], a_t, s_{t+1})b_t(\mathcal{H}_j)}^{P(A)}}{\underbrace{Pr(o_{t+1} | b_t, a_t)}_{P(B)}},
\end{aligned}
\label{Eq_Belief}
\end{equation}
where $b_t(\mathcal{H}_i)$ is the previous belief (probability) that the occupant is human model $\mathcal{H}_i$, $b_{t+1}(\mathcal{H}_i)$ is the updated belief for the next time step, and $Pr(o_{t+1} |b_t, a_t)$ is the probability of observing $o_{t+1}$ when action $a_t$ is taken with the belief vector $b_t$ over all possible human models (or occupants) $\mathcal{H}$. The denominator is given by:
\begin{equation}
\begin{aligned}
    & Pr(o_{t+1} | b_t, a_t) =  \sum_{\mathcal{H}_k\in\mathcal{H}}O(a_t,  o_{t+1}, [o_{t+1}, \mathcal{H}_k]) & \\
    & \times \sum_{\mathcal{H}_j \in \mathcal{H}} T([o_t, \mathcal{H}_j], a_t, s_{t+1})b_t(\mathcal{H}_j).
\end{aligned}
\end{equation}
It should be noted that in the context of our problem, the transition function $T$ and observation function $O$ are unknown. Therefore, in the next sections we will describe our solution around this issue.

\subsection{Partially Observable Smart Home System}
Here, we describe a modified SHS that implements a Bayesian model to learn multiple user's personalized TH preferences when the user information is limited. Accordingly, we define a Partially Observable Smart Home System (POSHS) as a Bayesian model \cite{POMDP_Bayes, POMDP_Bayes_2} of the SHS where the state of the environment is not fully observable as discussed in the previous section.
We will further assume that the model of the occupant does not change during an episode (a set of 3 activities). We will then use the Bayes theorem and the learnt occupants' TH preferences to infer the current human model in the environment, thus identifying it.

\begin{figure*}[t]
    \centering
    \subfloat[POSHS during the episode]{
    \includegraphics[width=0.8\columnwidth]{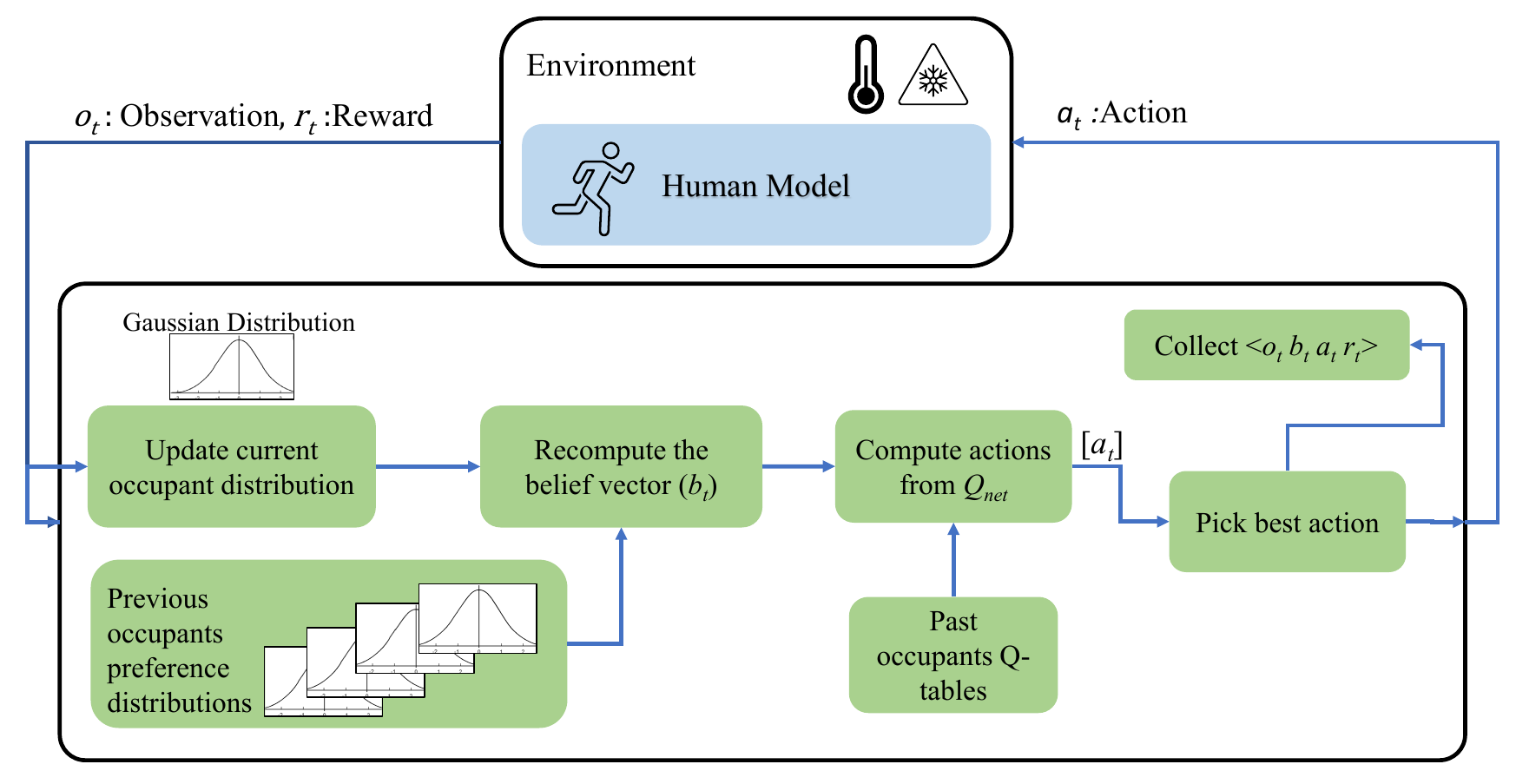}
    }
    \par
    \subfloat[POSHS at the end of episode]{
    \includegraphics[width=0.77\columnwidth]{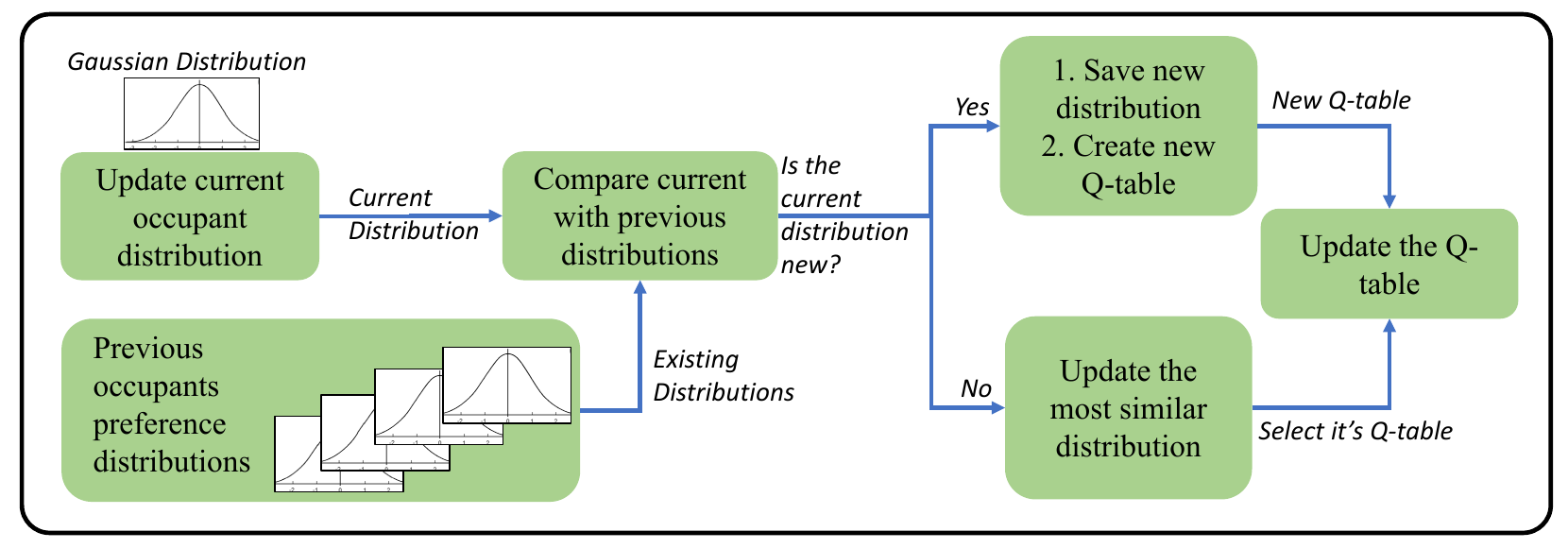}
    }
    \caption{(A) POSHS during the episode: belief is computed and action is taken based on the current belief vector. The observation transitions are stored and distribution parameters are updated. (b) POSHS at the end of the episode: the current observed distribution is compared with stored distributions. If the distance between the distributions is greater than a threshold, a new $Q$-table is created for the new human occupant, otherwise the parameters of the distribution most similar to the current distribution are updated.}
    \label{Fig_POSHS}
\end{figure*}

We depict an overview of POSHS in Figure \ref{Fig_POSHS}, where it is divided into two components (panels (a) and (b)). During the episode (panel (a)), at each time step, the agent  estimates a probability distribution over the TH preferences of the current occupant based on the TH and activity observations. This distribution is combined with a pool of previously learned user TH preference distributions using the Bayes theorem to estimate the belief state over the possible current occupants. The belief state is then used to weight the $Q$-table of each possible occupant in order to select the optimal action for the current belief. At the end of the episode (panel (b)), the agent compares the final estimated TH preference distribution to the pool of learned preferences. If a match is found, then the episode distribution parameters are used to update that occupant's TH preference distribution. If a match is not found, we assume this is a new user. Therefore, the estimated probability TH preference distribution is added to the pool along with a new $Q$-table. Finally, the Q-tables are updated with the episode data. In the next subsections, we describe these elements (TH preference distribution, belief estimation, distribution similarity, and Q update) in depth.


\subsection{TH Preference Distribution}
 
Since it is impossible to fully define $T$ and $O$ for an unknown set of occupants, we need a different approach to estimate the SHS belief over the set of possible occupants during the episode. To make the problem more tractable, we first assume that the occupants are the same through the whole episode. 
This implies that the transitional probabilities in an episode have the following property:
\begin{equation}
\begin{split}
    T([o_t, \mathcal{H}_i], a_t, [o_{t+1}, \mathcal{H}_j]) &= 0  \quad \forall i \neq j\\
    T([o_t, \mathcal{H}_i], a_t, [o_{t+1}, \mathcal{H}_i]) &= 1 .
\end{split}
\end{equation}

Accordingly, given a potential occupant $\mathcal{H}_i$, we need to define a probability distribution over the TH preference for each activity ($P(B|A)$ in Eq. \ref{Eq_Belief}). Using the Bayes rule, this will provide a way to estimate the probability of a specific occupant, given the TH observation ($P(A|B)$ in Eq. \ref{Eq_Belief}). The TH preference probability distribution $P(TH|\mathcal{H}_i)$ of a given occupant $\mathcal{H}_i$ over temperature or humidity for each activity is described by a Gaussian distribution given as:
\begin{equation}
    P(k | \mathcal{H}_i) = \frac{1}{\sigma_{t}\sqrt{2\pi}} 
    \left.\exp\left( -\frac{1}{2}\left(\frac{k-\mu_{\mathcal{H}_i}}{\sigma_{\mathcal{H}_i}}\right)^{\!2}\,\right)\right\vert_{k = T \text{ or }H}
    \label{Eq_PDF}
\end{equation}
where $k$ is an observed scalar value of temperature (T) or humidity (H) and $\mu_{\mathcal{H}_i}$ and $\sigma_{\mathcal{H}_i}$ are the mean and standard deviations of the of the preference distribution for occupant $\mathcal{H}_i$. \textit{The above equation is defined individually for both temperature and humidity, and for each activity.} This Gaussian distribution allows the system to support individual preference variabilities represented by $\mu$ and $\sigma$ as the mean and standard deviation of the comfortable thermal values. It should be noted that each human model has $3$ distributions representing each given activity respectively. For convenience, we will be using the notation `TH' for the equations implying that they apply to both temperature and humidity.

Given a complete episode, the SHS samples each observation when the human agent is not making any changes to the TH and continues with a given activity. Accordingly, the parameters of the TH preference distribution for the occupant of that episode can be estimated using best point estimators given by
\begin{equation}
    \mu_e = \frac{1}{C}\sum_{\text{valid } t=0}^{T} TH_t
    \label{Eq_Mean}
\end{equation}
and
\begin{equation}
    \sigma_e = \sqrt{\frac{1}{C}\sum_{i=0}^{T}(TH_i-\mu_t)^2}, 
    \label{Eq_Standard_Deviation}
\end{equation}
where $TH_t$ is the observed temperature or humidity at time step $t$ for the given activity, $C$ is the number of valid observations, and $T$ is the length of the episode. 
Therefore, a complete TH preference probability distribution for a model can be coded using a 12d vector (2 parameters $\times$ 2 TH $\times$ 3).

\subsection{Belief Estimation}
Here, we describe the mathematical formulation of the belief estimation of the smart home about the human occupant at time step $t$ given observation $TH_t$. For simplicity, let us assume that we have two human agents in a given smart home environment denoted by $\mathcal{H}_a$ and $\mathcal{H}_b$. Restructuring Eq. \ref{Eq_Belief} for human model $\mathcal{H}_a$ and $\mathcal{H}_b$ based on the Bayes theorem, we obtain:
\begin{equation}
    \frac{P(TH_t|\mathcal{H}_a)}{P(TH_t|\mathcal{H}_b)} = \frac{P(\mathcal{H}_a|TH_t) P(\mathcal{H}_b)}{P(\mathcal{H}_b|TH_t) P(\mathcal{H}_a)}. \\
\label{P_THA_by_P_THB}
\end{equation}
Solving the above equation for $\mathcal{H}_a$ and $\mathcal{H}_b$, we obtain:
\begin{equation}
    P(\mathcal{H}_a|TH_t) = \frac{P(TH_t | \mathcal{H}_a)P(\mathcal{H}_a)}{P(TH_t|\mathcal{H}_a)P(\mathcal{H}_a)+ P(TH_t|\mathcal{H}_b)P(\mathcal{H}_b)} 
\label{Eq_Ha_Posterior}
\end{equation}
and 
\begin{equation}
    P(\mathcal{H}_b|TH_t) = \frac{P(TH_t | \mathcal{H}_b)P(\mathcal{H}_b)}{P(TH_t|\mathcal{H}_a)P(\mathcal{H}_a)+ P(TH_t|\mathcal{H}_b)P(\mathcal{H}_b)}. 
\label{Eq_Hb_Posterior}
\end{equation}

Given $N$ different human models, we can define the probability of the $i^{th}$ human model $\mathcal{H}_i$ at the $t_{th}$ time-step as:
\begin{equation}
    P(\mathcal{H}_i|TH_t) = \frac{P(TH_t|\mathcal{H}_i)P(\mathcal{H}_i)}{\sum_{i = 0}^{N} P(TH_t|\mathcal{H}_i)P(\mathcal{H}_i)}. 
\end{equation}
Dividing $P(\mathcal{H}_a|TH_t)$ by $P(\mathcal{H}_b|TH_t)$ from Eq. \ref{Eq_Ha_Posterior} and Eq. \ref{Eq_Hb_Posterior}, and then substituting $P(TH_t|\mathcal{H})$ with Eq. \ref{Eq_PDF} we obtain:
\begin{equation}
\small
\begin{split}
    \frac{P(\mathcal{H}_a|TH_t)}{P(\mathcal{H}_b|TH_t)}  = \frac{\frac{1}{\sigma_{\mathcal{H}_a}\sqrt{2\pi}} 
    \exp\left( -\frac{1}{2}\left(\frac{TH_t-\mu_{\mathcal{H}_a}}{\sigma_{\mathcal{H}_a}}\right)^{2}\,\right)P(\mathcal{H}_a)}{\frac{1}{\sigma_{\mathcal{H}_b}\sqrt{2\pi}} 
    \exp\left( -\frac{1}{2}\left(\frac{TH_t-\mu_{\mathcal{H}_a}}{\sigma_{\mathcal{H}_a}}\right)^{2}\,\right)P(\mathcal{H}_b)} \\
    = \frac{\sigma_{\mathcal{H}_b}}{\sigma_{\mathcal{H}_a}}\frac{
    \exp\frac{1}{2}\left(\left(\frac{TH_t-\mu_{\mathcal{H}_a}}{\sigma_{\mathcal{H}_a}}\right)^{\!2} - \left(\frac{TH_t-\mu_{\mathcal{H}_b}}{\sigma_{\mathcal{H}_b}}\right)^{\!2}\,\right)P(\mathcal{H}_a)}{P(\mathcal{H}_b)} .
\end{split}
\label{Equ_Pa/Pb}
\end{equation}
Assuming $\sigma_{\mathcal{H}_a} \approx \sigma_{\mathcal{H}_b} = \sigma$, we get:
\begin{equation}
\begin{aligned}[b]
    & \frac{P(\mathcal{H}_a|TH_t)}{P(\mathcal{H}_b|TH_t)} = 
    \frac{
    \exp\frac{1}{2\sigma^2}\left(\left(2TH_t-\mu_{\mathcal{H}_a} - \mu_{\mathcal{H}_b}\right)\,\right)}{\frac{P(\mathcal{H}_b)}{\left(\mu_{\mathcal{H}_a} - \mu_{\mathcal{H}_b} \right)P(\mathcal{H}_a)}} .
\end{aligned}
    \label{Eq_P_A_by_P_B}
\end{equation}

At the beginning of the training episode, the POSHS agent has no information about the user, thus maintains an equal initial belief for each possible human occupant. Accordingly, the prior probability of each model can be defined as $P(\mathcal{H}_a) = P(\mathcal{H}_b)$. Hence, we can rewrite Eq. \ref{Eq_P_A_by_P_B} as 
\begin{equation}
\begin{aligned}
    & \frac{P(\mathcal{H}_a|TH)}{P(\mathcal{H}_b|TH)} + 1 =
    \frac{P(\mathcal{H}_a|TH) +  P(\mathcal{H}_b|TH)}{P(\mathcal{H}_b|TH)} =  \frac{1}{P(\mathcal{H}_b|TH)} \\
    & = \exp\frac{1}{2\sigma^{2}}\left(\left(2TH_t-\mu_{\mathcal{H}_a} - \mu_{\mathcal{H}_b}\right)\left(\mu_{\mathcal{H}_a} - \mu_{\mathcal{H}_b} \right)\,\right) + 1 .
\end{aligned}
\end{equation}
This gives 
\begin{equation}
\begin{aligned}
    & P(\mathcal{H}_b|TH) = \\
    & \frac{1}{\exp\frac{1}{2\sigma^{2}}\left(\left(2TH_t-\mu_{\mathcal{H}_a} - \mu_{\mathcal{H}_b}\right)\left(\mu_{\mathcal{H}_a} - \mu_{\mathcal{H}_b} \right)\,\right) + 1}
\end{aligned}
\end{equation}
and 
\begin{equation}
\begin{aligned}
    & P(\mathcal{H}_a|TH_t) = \\ & \frac{\exp\frac{1}{2\sigma^{2}}\left(\left(2TH_t-\mu_{\mathcal{H}_a} - \mu_{\mathcal{H}_b}\right)\left(\mu_{\mathcal{H}_a} - \mu_{\mathcal{H}_b} \right)\,\right)}
    {\exp\frac{1}{2\sigma^{2}}\left(\left(2TH_t-\mu_{\mathcal{H}_a} - \mu_{\mathcal{H}_b}\right)\left(\mu_{\mathcal{H}_a} - \mu_{\mathcal{H}_b} \right)\,\right) + 1}.
    \end{aligned}
\end{equation}

With the above equation, we derive the \textit{initial} posterior for each human agent at each time step for two human occupants. At the next time step we perform the Bayesian update where the posterior ($P(\mathcal{H}_i|HT_t)$) at time step $t$ becomes the prior ($P(\mathcal{H}_i)$) at time step $t+1$. 
Accordingly, we can take Eq. \ref{P_THA_by_P_THB} and divide both numerator and denominator by the $P(TH_t|\mathcal{H}_a)$, and then replace the ratio in the second term of the denominator by Eq. \ref{Equ_Pa/Pb}. This yields:
\begin{equation}
\begin{aligned}
   & P(\mathcal{H}_a|TH_t)  = 
   \frac{P(TH_t |     \mathcal{H}_a)P(\mathcal{H}_a)}{P(TH_t|\mathcal{H}_a)P(\mathcal{H}_a)+ P(TH_t|\mathcal{H}_b)P(\mathcal{H}_b)}\\
   & = \frac{
    P(\mathcal{H}_a)}{P(\mathcal{H}_a) + 
    \frac{P(TH_t|\mathcal{H}_b)}{P(TH_t|\mathcal{H}_a)}P(\mathcal{H}_b)}\\
    & = \frac{
    P(\mathcal{H}_a)}{P(\mathcal{H}_a) + 
    C(\mathcal{H}_b, \mathcal{H}_a)P(\mathcal{H}_b)} .
\end{aligned}    
\end{equation}
where 
\begin{equation}
    C(\mathcal{H}_b, \mathcal{H}_a) = \exp \Big(\frac{\left(2 TH_t-\mu_{\mathcal{H}_b} - \mu_{\mathcal{H}_a}\right)\left(\mu_{\mathcal{H}_b} - \mu_{\mathcal{H}_a} \right)\,\Big)}{2\sigma^{2}}.
\end{equation}

Similarly, for Model $\mathcal{H}_b$, the posterior can be computed as:
\begin{equation}
\begin{aligned}
    & P(\mathcal{H}_b|TH) = \\
    & \frac{P(\mathcal{H}_b)}{
    C(\mathcal{H}_a, \mathcal{H}_b)P(\mathcal{H}_a) + P(\mathcal{H}_b)} .
    \end{aligned}
\end{equation}
Expanding the above equations for $N$ human models, we can get our posterior for the $i^{th}$ human model as:
\begin{equation}
    P(\mathcal{H}_i|TH_t) = 
    \frac{P(\mathcal{H}_i)}{
    \sum_{j=0}^{N} C(\mathcal{H}_j, \mathcal{H}_i)P(\mathcal{H}_j)} ,
\end{equation}

\subsection{Distribution Similarity}
Upon completion of an episode, the SHS agent needs to know if the current occupant belongs to the pool of previously observed occupants or if it is a new occupant in the environment. To do so, the SHS agent follows the procedure in Figure \ref{Fig_POSHS}(b). Here, the best point estimators ($\mu_e$ and $\sigma_e$ from Eqs. \ref{Eq_Mean} and \ref{Eq_Standard_Deviation}) for the current occupant TH distribution are computed. The resulting TH preference distribution is then compared with the distributions in the pool of previously observed occupants using the Jensen-Shannon Divergence (JSD) measure \cite{JSD}. We select this particular divergence measurement since it has shown success in similar RL-related applications in prior works \cite{Rene_JSD}. Moreover, this metric provides benefits over other divergence metrics, in particular, 
(\textit{a}) when variations between distributions are small, the divergence remains smooth,
and (\textit{b}) the divergence between distributions is symmetric, i.e., $JSD(P_0, P_1) = JSD(P_1, P_0)$.

The Jensen-Shannon divergence between $n$ probability distributions is formulated as:
\begin{equation}\label{JSD}
    JSD = H\left[\sum_{i=1}^{n} w_ip_i\right] - \sum_{i=1}^{n}w_iH(p_i), 
\end{equation}
where $H$ is the Shannon entropy \cite{Entropy_Shannon} and $w_i$ are the weights selected for each probability distribution $p_i$. 
For our problem, we compare only two distributions at an instance, i.e., the current distribution of the occupant in question and each of the distributions from the pool of observed occupants. As a result, we set $n=2$, and thus $w_1 = w_2 = 0.5$. Next, we define a new term $\tau_{JSD}$ as the threshold such that
\begin{equation}
    JSD \geq \tau_{JSD} \iff \text{Different Human Model},
\end{equation} 
where $JSD$ is the measured Jensen-Shannon divergence between the two TH distributions. If the divergence between the distributions is greater than $\tau_{JSD}$, the current occupant's distribution is added to the pool and the SHS agent creates a new $Q$-table. Otherwise the distribution with the smallest divergence (distribution with highest likelihood) is updated using a moving average with the current estimated parameters to improve the long term estimation of the occupant's preference distribution.


\subsection{Q Update}
Here, we describe the computation of the $Q$ values of state-action pairs for a given occupant, followed by optimally selecting the best action using the belief vector over the user space. During each episode as shown in Figure \ref{Fig_POSHS}(a), the SHS agent updates the TH distribution estimation parameters. It then calculates the belief vector using Bayes rule after which the $Q$-tables of the observed occupants are weighted using their belief vector to get a summarized $Q_{net}$ value. The SHS agent then chooses the optimal action from the $Q_{net}$ values of all the possible actions, which is then executed in the environment. The observation $o_t$, computed belief $b_t$, action taken $a_t$, and the reward received $r_{t+1}$ from the environment are all stored in the memory.

To calculate the weighted $Q$ update, each weight is decided by the belief of the specific human model for a given state from the belief vector. This update is given by the following equation:
\begin{equation}\label{Q_Update}
    Q_{net}(o_t, a_t) = \sum^{\mathcal{H}} b_t(\mathcal{H})Q_{\mathcal{H}}(o_t, a_t) ,
\end{equation}
where $o_t$ is the observation, $a_t$ is the chosen action, the weight $b_t(\mathcal{H}) = P(\mathcal{H}|TH_t)$ is the conditional probability given the current \textit{TH}, and $Q_{net}$ is the weighted sum of all $Q$ values of all the occupants for a given observation.

Accordingly, we now update the $Q$ table for a given occupant $\mathcal{H}$. Let $b_t$ be the belief vector at time step $t$ in the episode and $b_t(\mathcal{H})$ be the belief component of $b_t$ for human $\mathcal{H}$. Moreover, let $o_t$ be the received observation at time step $t$ in the episode, $\alpha = 0.05$ be the learning rate, $\gamma = 0.98$ be the discount factor, and $r_{t+1}$ be the reward received after action $a_t$ from observation $o_t$ at time step $t$. Accordingly, the $Q_{\mathcal{H}}$ update is given by
\begin{equation}
\begin{split}
    Q_{\mathcal{H}}(o_t, a_t) = (1 - \alpha)Q_{\mathcal{H}}(o_t, a_t) + \alpha[r_{t+1}*b_t(\mathcal{H}) + \\ \gamma \underset{a'}{max} Q_{\mathcal{H}}(o_{t+1}, a')].
\end{split}
\end{equation}
We obtain $\alpha = 0.05$ and $\gamma = 0.98$ with some preliminary experiments such that the model converges.
With the updated $Q_{net}$ value, the agent can select the optimal action. For a given observation $o_t$, actions are selected using a decaying $\epsilon$\nobreakdash-Greedy policy keeping a balance between exploration and exploitation. The policy is given by:
\begin{equation}
\pi(o_t) = \begin{cases}
    \underset{a\in\mathcal{A}}{\argmax}\{Q_{net}(o_t, a)\} \text{with probability } 1-\epsilon\\
    a\in\mathcal{A} \text{ with probability } \epsilon/|\mathcal{A}|
\end{cases},
\end{equation}
where $\epsilon$ is a threshold value less than 1 that decays exponentially down to $0.005$, and $\mathcal{A}$ is the set of all possible actions.

\subsection{Pseudocodes}
In order to better facilitate reproduction of our work, in this subsection we present the pseudocodes and detailed algorithms for computing the Jensen-Shanon divergence, action selection and belief estimation, $Q$ table update, and POSHS, in Algorithms \ref{Algo_JSD}
through
\ref{Algo_POSHS}.

\begin{algorithm}
\caption{Jensen-Shannon Divergence}\label{Algo_JSD}
\begin{algorithmic}[1]
    \Function{JSD}{Distribution $d_a$, Distribution $d_b$, Amplification Factor $A_f$}
    \State let $d_c = $ $\frac{d_a+d_b}{2}$
    \State let $H_a = d_a \times log{\frac{d_a}{d_c}}$ \Comment{Calculate Shannon entropy for $d_a$}
    \State let $H_b = d_b \times log{\frac{d_b}{d_c}}$ \Comment{Calculate Shannon entropy for $d_b$}
    \State $\tau$ = $\sqrt{\frac{\sum H_a + \sum H_b}{2}}$
    \State return $A_f * \tau$
    \EndFunction
\end{algorithmic}
\end{algorithm}

\begin{algorithm}
\caption{GetAction}\label{Algo_GetAction_POSHS}
\begin{algorithmic}[1]
    \Function{getAction}{$o_t$}
    \State let $N = $ number of human models
    \State let $TH_t \gets $ $o_t$ 
    \State calculate belief $P(\mathcal{H}_i|TH_t) = \frac{P(TH_t|\mathcal{H}_i)P(\mathcal{H}_i)}{\sum_{i = 0}^{N} P(TH_t|\mathcal{H}_i)P(\mathcal{H}_i)}$ \Comment{Belief update}
    \State $Q_{net}(o_t, a_t) = P(\mathcal{H}_a)Q_a(o_t, a_t) + P(\mathcal{H}_b)Q_b(o_t, a_t) + \hdots + P(\mathcal{H}_n)Q_n(o_t, a_t)$
    \State \textbf{return} $\underset{a}{\argmax} Q(o_t, a)$
    \EndFunction
\end{algorithmic}
\end{algorithm}

\begin{algorithm}
\caption{Update}\label{Algo_Update_POSHS}
\begin{algorithmic}[1]
    \Function{update}{$memory, newNode$}
    \State let $N = $ number of human models
    \For{$\mathcal{H}$ in all human models}
        \State let $b = P(\mathcal{H})$   \Comment{Belief $i_{th}$ human model}
        \If{newNode}
            \State $Q_{\mathcal{H}} \gets 0$ \Comment{initialize $Q$ table for unobserved human model}
        \EndIf
        \State \textbf{Endif}
        \For{$<o_t, a_t, o_{t+1}, r_{t+1}>$ in $memory$}
            \State $Q_{\mathcal{H}}(o_t, a_t) = (1 - \alpha)Q_{\mathcal{H}}(o_t, a_t) + \alpha[r_{t+1}*b + \gamma \underset{a'}{max} Q_{\mathcal{H}}(o_{t+1}, a')]$
        \EndFor
        \State \textbf{Endfor}
    \EndFor
    \State \textbf{Endfor}
    \EndFunction
\end{algorithmic}
\end{algorithm}

\begin{algorithm}
\caption{Partially Observable Smart Home System (POSHS)}\label{Algo_POSHS}
\begin{algorithmic}[1]
    \State let $\mu_{TH} = $ \{ \} \Comment{Set containing Model's mean TH}
    \State let $\sigma_{TH} = $ \{ \} \Comment{Set containing Model's deviation of TH}
    \State let $Q_{\mathcal{H}} = $ \{ \} \Comment{Set containing Model's Q Table}
    \State let $o_0 = \mathcal{O}(s_0)$ where $o_t$ is partial observation of state $s_t$
    \For{each $e$ in $episodes$}
        \State let $C = 0$ 
        \State let $\mu_e = 0$ \Comment{Current Episode mean of TH}
        \State let $\sigma_e = 0$ \Comment{Current Episode standard deviation of TH}
        \State let $memory =$ [ ]
        \While {$o_t$ is not terminal}
        \State with greedy policy $a_t = getAction(o_t)$ \Comment{Algorithm \ref{Algo_GetAction_POSHS}}
        \State execute $a_t$
        \State observe $o_{t + 1}, r_{t+1}$
        \State update $\mu_e, \sigma_e$ for $C$ time-steps
        \State append $<o_t, a_t, o_{t+1}, r_{t+1}>$ in $memory$
        \State $C = C$ + $1$
    \EndWhile
    \State \textbf{Endwhile}
    \State let $ newNode = False$ \Comment{Flag to indicate a new human user}
    \State let $ divergence = $ [ ]
    \For{$<\mu, \sigma>$ in $\mu_{TH}, \sigma_{TH}$}
        \State let $d_a = $ $PDF(\mu_e, \sigma_e$)
        \State let $d_b = $ $PDF(\mu, \sigma$)
        \State let $k = $ $JSD(d_a, d_b)$ \Comment{Algorithm \ref{Algo_JSD}}
        \State append $k$ in $divergence$
    \EndFor
    \State \textbf{Endfor}
    \If{$min(distance) > \tau_{JSD} \And !newNode$}
        \State add $\mu_e, \sigma_e$ in $\mu_{TH}, \sigma_{TH}$
        \Comment{Add New Human Node}
        \State $newNode = True$
    \ElsIf{$!newNode$}
        \State let $i = \argmin(divergence)$
        \State let $m = 0.5$ \Comment{Moving filter significance}
        \State $\mu_{TH}[ i ] = (1-m)\mu_e + m\mu_{TH}[i]$
        \State $\sigma_{TH}[ i ] = (1-m)\sigma_e + m\sigma_{TH}[i]$
    \EndIf
    \State \textbf{Endif}
    \State $update(memory, newNode)$ \Comment{Algorithm \ref{Algo_Update_POSHS}}
    \EndFor
    \State \textbf{Endfor}
\end{algorithmic}
\end{algorithm}

,

\section{Baseline}
We compare our model (POSHS) against two baselines: (1) a Recurrent Neural Network (RNN), and (2) a Transformer Attention layer. In this section we describe the details of these two models.

Classical RNNs, however, often suffer from issues such as vanishing gradient, making it difficult for the model to remember long-term information. As a result, we use a Long Short-Term Memory (LSTM) networks as our baseline. LSTM is a type of RNN that can hold a sequence of information in its memory and learn temporal relations. To do this, LSTM uses long-term sequences in memory referred to as `cell-state'. The output from previous step are referred to as `hidden-state'.
Information in an LSTM is controlled via three gates given as (\textit{a}) forget-gate, which decides the information that needs to be rejected or accepted, (\textit{b}) input-gate, which is used to control the information into the cell-state, and (\textit{c}) output-gate, which generates the output for each time-step.

In a POMDP environment, Deep Q Network (DQN) fails to learn the optimal policy because it assume complete information about the state which we lack in our problem. Unlike DQNs, LSTM networks are capable of remembering long sequences of TH preferences and encoding the observation onto a latent space vector, making it possible to approximate the underlying hidden state \cite{POMDP_LSTM, POMDP_LSTM_2}.
Thus, unlike POSHS where the agent needs to recognize the hidden state of the human occupant using Bayes rule explicitly, LSTMs can implicitly learn the underlying states using its recurrent memory. 

\begin{figure}[t]
    \centering
    \includegraphics[width=0.6\columnwidth]{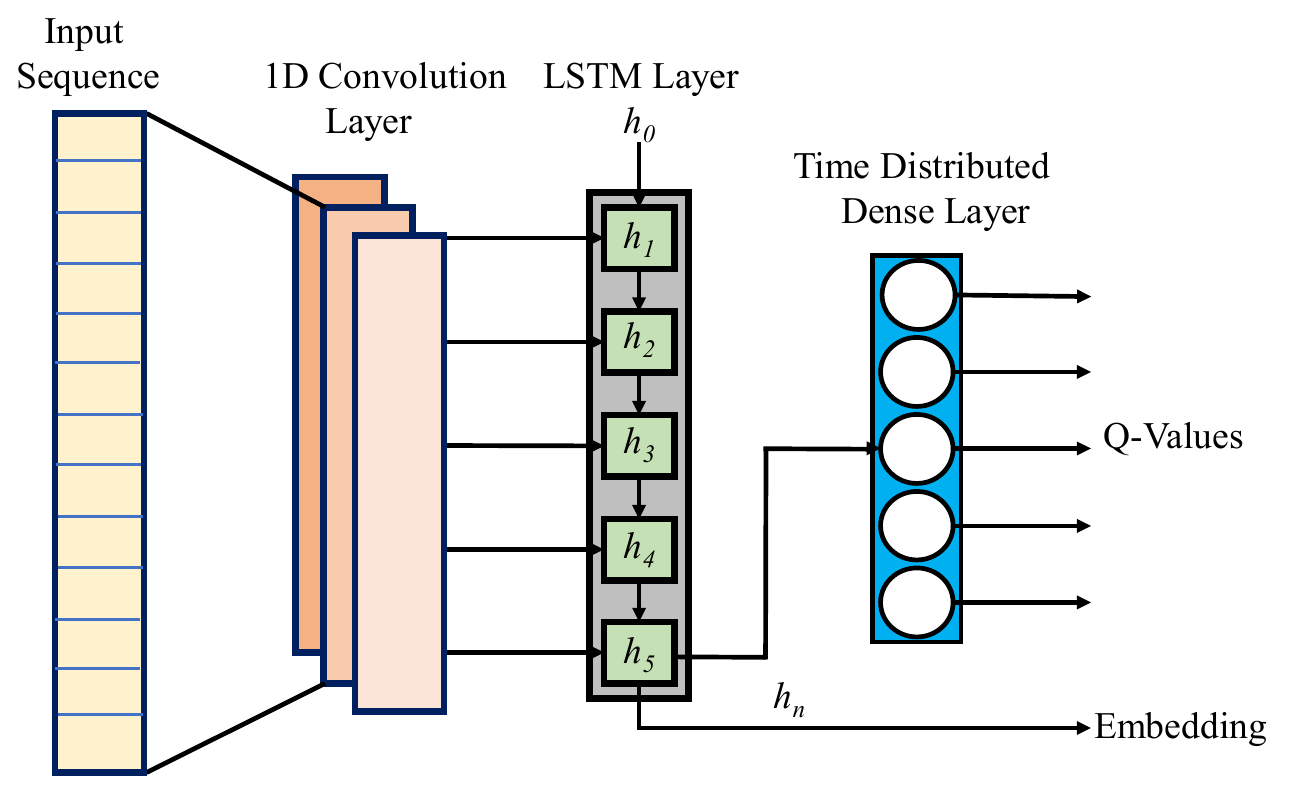}
    \caption{LSTM baseline architecture: the input is a variable length sequence with 5d features which is convoluted in 1-dimension. The convoluted outputs is fed to the LSTM layer which is connected with a Time Distributed dense layer outputting five $Q$-values for five actions to be taken by the SHS.}
    \label{fig_RNN_SHS_Architecture}
\end{figure}

To use LSTM as a baseline, we design two networks namely \textit{train} and \textit{target}. Both networks have the same architecture and are initialized with the same weights and biases. The inputs are sequences of TH, activity, and human action of the occupant. The input layer of the model is a 1d convolutional layer with 32 filters, a kernel size of 1, a stride of 1, and a Rectified Linear Unit (ReLU) activation function. The output of this layer is fed to an LSTM layer with 175 cells followed by a tanh activation function. In the end, the output of this layer is fed to a dense layer of 5 neurons and ReLU activation to output the $Q$-values for each possible action (total of 5 actions: increase/decrease T and H, and no action). We set the learning rate as $0.0013$ obtained empirically to keep the networks stable. Like the POSHS model, the $\epsilon$ decays exponentially down to 0.005. We do not decrease it completely to zero to explore all possible states.
The training procedure for this baseline is described in detail in Algorithm \ref{Algo_LSTM_SHS} with its architecture shown in Figure \ref{fig_RNN_SHS_Architecture}.

\begin{algorithm}[t]
\caption{Deep Reccurent Q-Learning for POMDP Environment}\label{Algo_LSTM_SHS}
\begin{algorithmic}[1]
    \State Initialize $train = $ train network with weights $\theta$
    \State Initialize $target = $target network with initial weights $ \theta^- = \theta$
    \State let $stateMemory = $ [ ]
    \State let $C = 0$ 
    \State let $o_t = \mathcal{O}(s_t)$ where $o_t$ is partial observation of state $s_t$ at time step $t$
    \While{$i < $ $episodes$}
    \While {$o_t$ is not terminal}
    \State with probability $\epsilon$ select random action 
    \State else select $a_t = $ $\underset{a}{\argmax} Q_t(o_t, h_{t-1}, a | \theta)$
    \State execute $a_t$
    \State observe $o_{t + 1}, r_t$
    \State Store the transition $<o_t, a_t, o_{t+1}, r_t>$ in $stateMemory$
    \State $C = C$ + $1$
    \EndWhile
    \State \textbf{Endwhile}
    \For{$<o_t, a_t, o_{t+1}, r_t>$ in $stateMemory$}
    \State $Q_{t} \gets$ Predict $Q | \theta$ for $o_{t+1}$ from $train$
    \State $Q^- \gets$ Predict $Q | \theta^-$ for $s_{t+1}$ from $target$
    \State $Q_{max_{t+1}}^-$ = $\max$ $Q^-$
    \If{$o_{t+1}$ is terminal}
        \State $y_t \gets r_t$
    \Else
        \State $y_t \gets r_t$ + $\gamma Q_{max_{t+1}}^{-}$
    \EndIf
    \State \textbf{Endif}
    \State calculate loss $\mathcal{L} = $ $(y_t - Q_t)^2$
    \EndFor
    \State \textbf{Endfor}
    \State fit $train$
    \State After $C$ steps $\theta^- \gets \theta$
    \EndWhile
    \State \textbf{Endwhile}
\end{algorithmic}
\end{algorithm}

\begin{figure}[t]
    \centering
    \hspace*{0.5cm}\includegraphics[width=0.6\columnwidth]{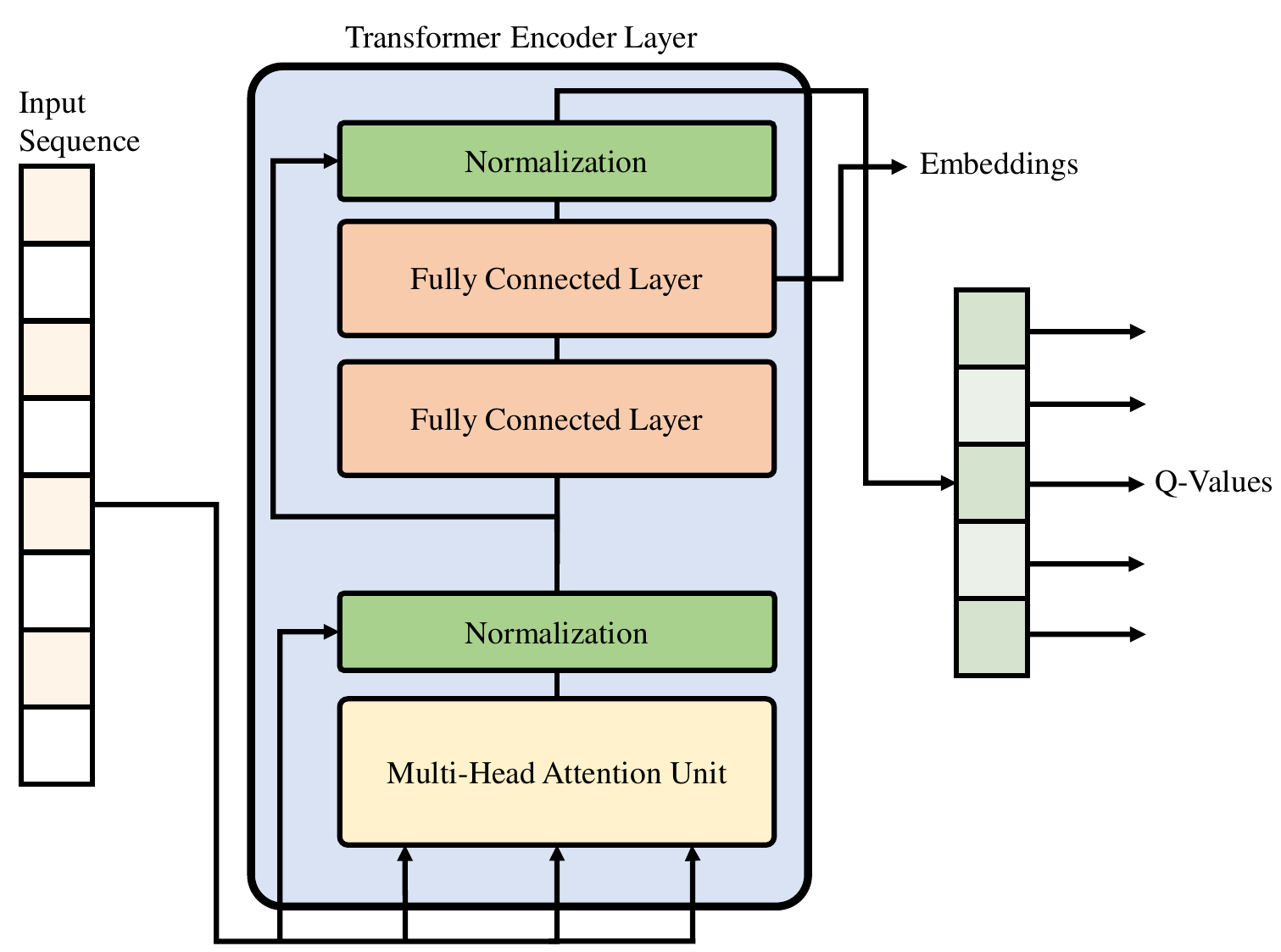}
    \caption{Transformer Attention baseline architecture: the input is a fixed length sequence with 5d features which is then passed to a multi-head attention layer. The output is fed to the normalization layer which is connected with a fully connected layer outputting five $Q$-values for five actions to be taken by the SHS.}
    \label{fig_Transformer_Architecture}
\end{figure}

In this algorithm, in contrast to POSHS where we update the $Q$-values directly, we update the parameters of the network with which we predict the next $Q$-value. The $Q$-values of the current state from the \textit{train} network are updated using the $Q$-values of the next state from the \textit{target} network. Using the same network (\textit{train}) for predicting the target $Q$-value would cause instability in the target $Q$-value. 
The weights for the \textit{train} and \textit{target} networks are represented by $\theta$ and $\theta^-$ respectively. Target weights $\theta^-$ are updated after every \textit{C} iterations with $\theta$, where $C$ is a hyper-parameter set to 7 empirically. 
We use Mean Squared Error as our loss function and ADAM \cite{ADAM} optimizer. We define our loss function similar to the update in table-based $Q$-learning which is the difference between the target and current $Q$-values. Accordingly, the loss function is given by
\begin{equation}
    \mathcal{L}(s_t, a_t | \theta) = (r_t + \gamma \underset{a}{max}Q(s_{t+1}, a | \theta^-) - Q(s_t, a_t| \theta))^2 .
\end{equation}

In order to recognize the user, the LSTM model uses the embedding of the sequence of TH preferences. However, for this embedding, the input sequence contains preferences of the occupant when it is \textbf{not} making any changes in TH which means that we only select TH preferences when the occupant is either executing the \textit{continue} or \textit{leave} action in a given activity. The embedding is obtained from the hidden state of the last cell from the LSTM layer of the architecture (Figure \ref{fig_RNN_SHS_Architecture}).

Before training, the initialized weight of the LSTM layer is same irrespective of the number of occupants. This helps the model to ensure that different embeddings are obtained for different TH sequences. The embedded information of TH sequence is collected at the end of episode when the model converges. At the end of episode, the obtained embedding is compared with each embedding stored in the pool.  To compare the embedding, we use Kolmogorov Smirnov Test (KST) \cite{KST}. This test quantifies the distance between two empirical distributions and returns the statistical distance and its $p$-value with which we can accept or reject the null hypothesis. For our problem, the null hypothesis is that two embeddings are same which means that both observed users are the same. The equation to compute KST for two emperical distributions $F_a$ and $F_b$ (of the embeddings of TH sequence) is given as:
\begin{equation}\label{KST}
    D, p = \underset{1\leq i\leq N}{max} \Big(F_a(y_i) - F_b(y_i)\Big)
\end{equation}
where $D$ is the statistical distance, $p$ is the $p$-value, $F(y)$ is the Cumulative Density Function (CDF) and $N$ is the number of samples. If $p$ is less than the significance level $\alpha$ then we can reject our null hypothesis which states that the two user embeddings are same. For our problem, after some preliminary tests we set the significance level $\alpha = 0.18$. We set the value higher than the general value of $0.05$ because the TH preference of occupants will always have some overlap. Thus discovering a new user is defined as:
\begin{equation}
    p \leq \alpha \iff \text{Different Human Model} 
\end{equation}

Using the above equations, we compare the similarity between the current occupant TH embedding and the embeddings of previously observed occupants. If the similarity is lower than $\alpha$, we add the embedding in the pool as a new occupant.
In order to identify the current user in the environment, we use the following equation:
\begin{equation}\label{Human_Select}
    \mathcal{H}_i = \underset{i}{\argmin} D_i
\end{equation}
where $D_i$ is the statistical distance between the current user's embedding and the $i^{th}$ embedding from the pool of previously observed users.

For the Transformer Attention \cite{transformer} baseline model, we follow a similar architecture to the LSTM model having a \textit{train} and \textit{target} network with the same number of inputs and $Q$-values as the action outputs. The input layer of this model is a linear layer followed by a transformer encoder layer with a three-head attention unit. This is followed by a ReLU activation function. Next, all the encoder layers are stacked and followed by a normalization layer. The normalized output is then passed to the fully connected layer that outputs the $Q$-values. Parameters like the learning rate and epsilon for this model were kept the same as that of the LSTM model. The loss between the current and the target $Q$-values is computed using the Mean Squared Error (MSE) which is then propagated backwards through the train network. This loss is computed between the train and target network's $Q$-values. These values are then optimized using ADAM optimizer. All the actions are taken using $Q$-values obtained from the train network. In order to avoid a race condition, after every $C$ steps, the weights of the target network are updated with the train network weights. Here $C$ is a hyper-parameter set to 7, which is found empirically to give a stable train network.

During training, it is possible for the model to recognize the user. We do this by using the encoder layers' stable outputs learned from the sequence provided as shown in Figure \ref{fig_Transformer_Architecture}. Similar to the LSTM model, we compare the obtained embeddings using KST as per Eq. \ref{KST}. After obtaining the KST distance, the identity is computed using Eq. \ref{Human_Select}. For this model, we set the significance value $\alpha=0.11$ obtained empirically to optimize the performance.

\section{Experiments and Results}
In this section, we present our results using the proposed POSHS and the baseline models (1) LSTM and (2) Transformer Attention layer. In the first experiment, we evaluate the POSHS performance by its accuracy in approximating the occupant's distribution accurately in the environment. In the second experiment, we evaluate the impact of the SHS on the occupant by measuring the time-steps required by the occupant to set TH with and without the SHS. Here, we refer time-steps as the number of steps the human model takes to set TH to an optimal comfort level. We further compare those results to the LSTM baseline.

\subsection{Experiment A} \label{Exp_A}
In this experiment, we measure the performance of the POSHS in identifying occupants while learning the thermal preferences of each user. 
We train each human model separately for 350 episodes such that each model can complete each activity while also setting the optimal TH. Next, we train the POSHS with the trained occupant for 150 episodes. During the training of the POSHS, a human model is chosen randomly before each episode.
Finally, each human model is run with the POSHS for 50 episodes to evaluate the final performance of the system. 



\begin{figure}[!t]
    \centering
    \includegraphics[width=0.6\columnwidth]{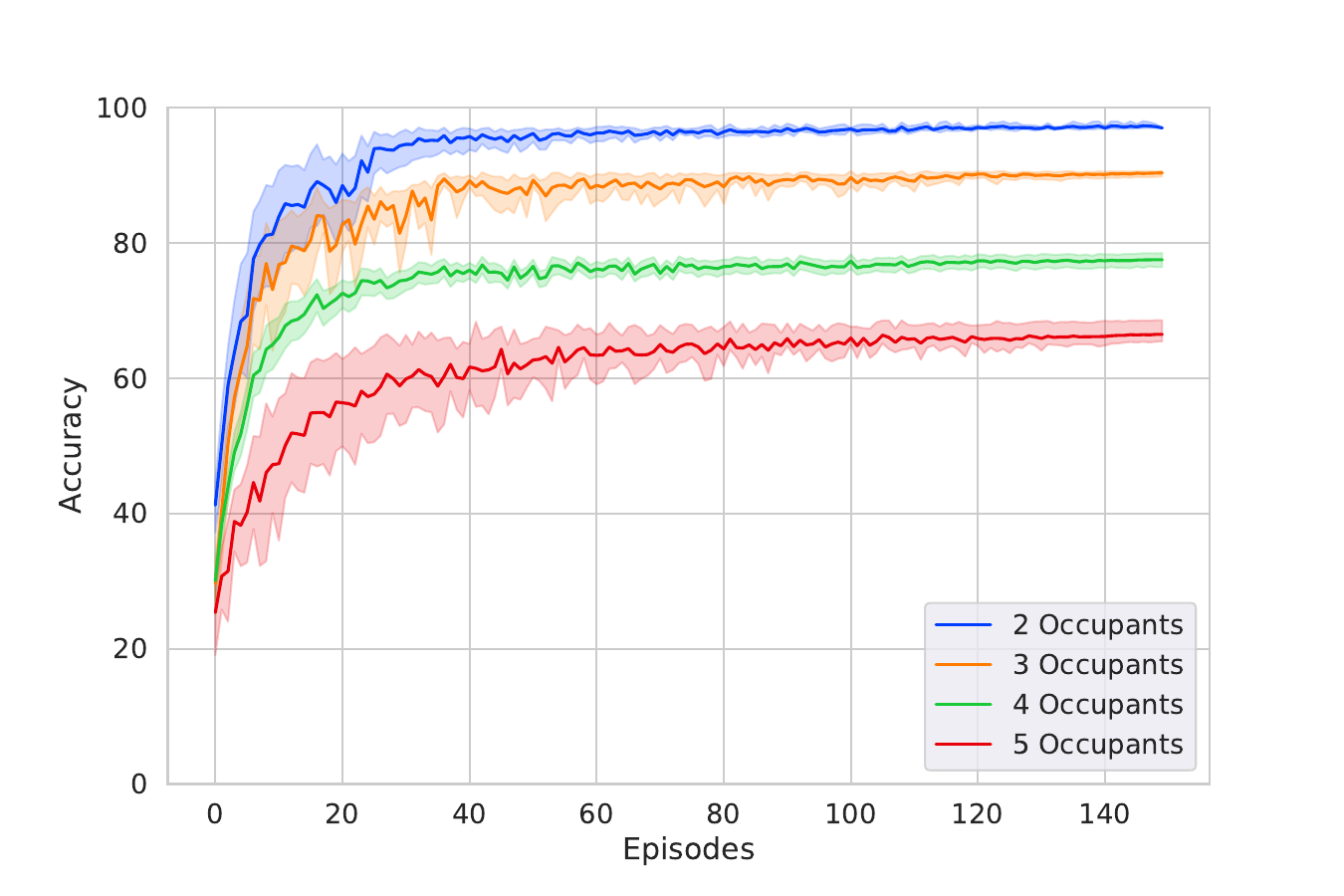}
    \caption{Training accuracy of 2-5 human models with POSHS agent using Jensen-Shannon Divergence to determine distribution similarity.}
   \label{Fig_Ha_Hb_Classification}
\end{figure}

First, we only experiment with two human models, namely $\mathcal{H}_a$ and $\mathcal{H}_b$. We set the metabolism indices for $\mathcal{H}_a$ to $[1, 1.2, 1.4]$ based on \cite{standard199255, cbe}. For $\mathcal{H}_b$, slightly different indices are required. We therefore add 0.05 to each of the indices for $\mathcal{H}_a$ to achieve $[1.15, 1.25, 1.45]$ for $\mathcal{H}_b$ for the PMV \cite{standard199255} range of $d_{0.25}$. We use $\tau_{JSD} = 0.13$ which empirically showed the best results in preliminary tests. Figure \ref{Fig_Ha_Hb_Classification} shows the accuracy of identifying the current model correctly during the training of POSHS. As discussed earlier, the occupant identification is performed by comparing the TH preference distribution estimated by the POSHS from the episode to the ones stored in the pool of distributions using the JSD, and selecting the best match. It can be seen on the Figure \ref{Fig_Ha_Hb_Classification} that the performance quickly achieves an accuracy of 85\% and converges to nearly 100\% accuracy. 



    
\begin{figure}[!t]
    \centering
    \subfloat[]{
    \includegraphics[width=0.5\columnwidth]{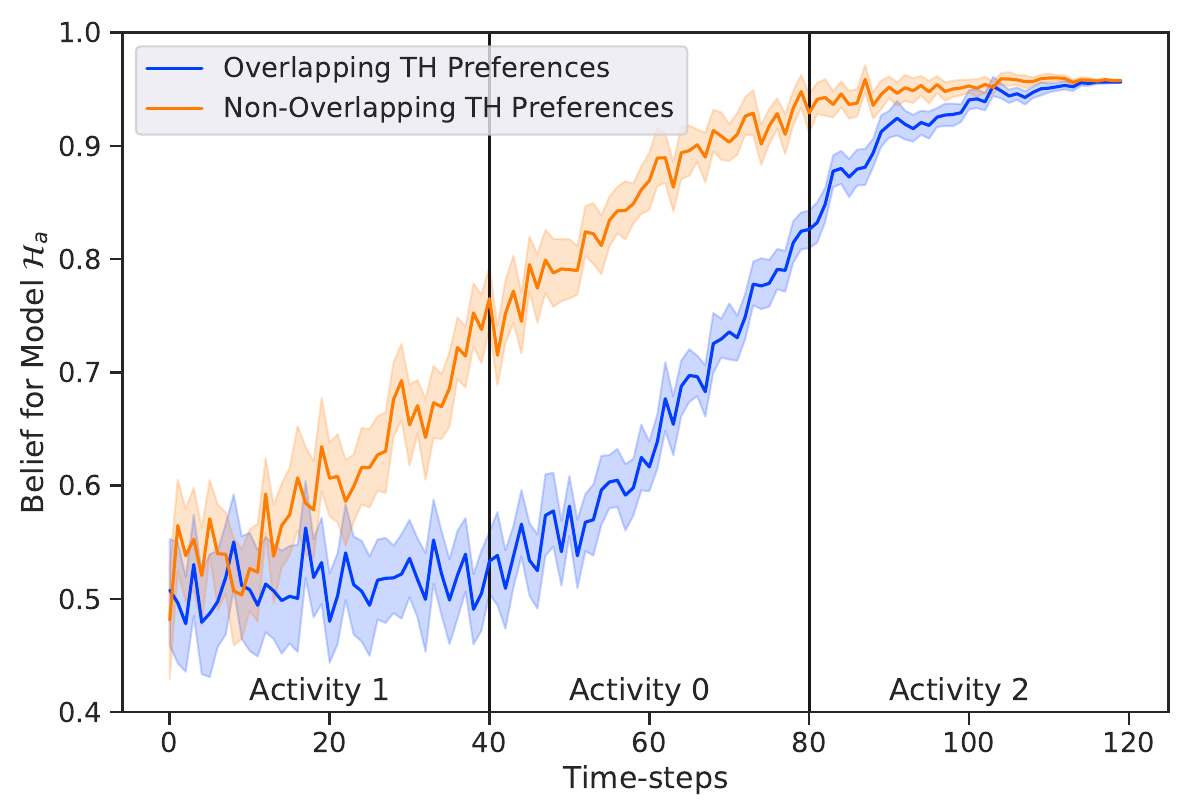}}
    \subfloat[]{
    \includegraphics[width=0.5\columnwidth]{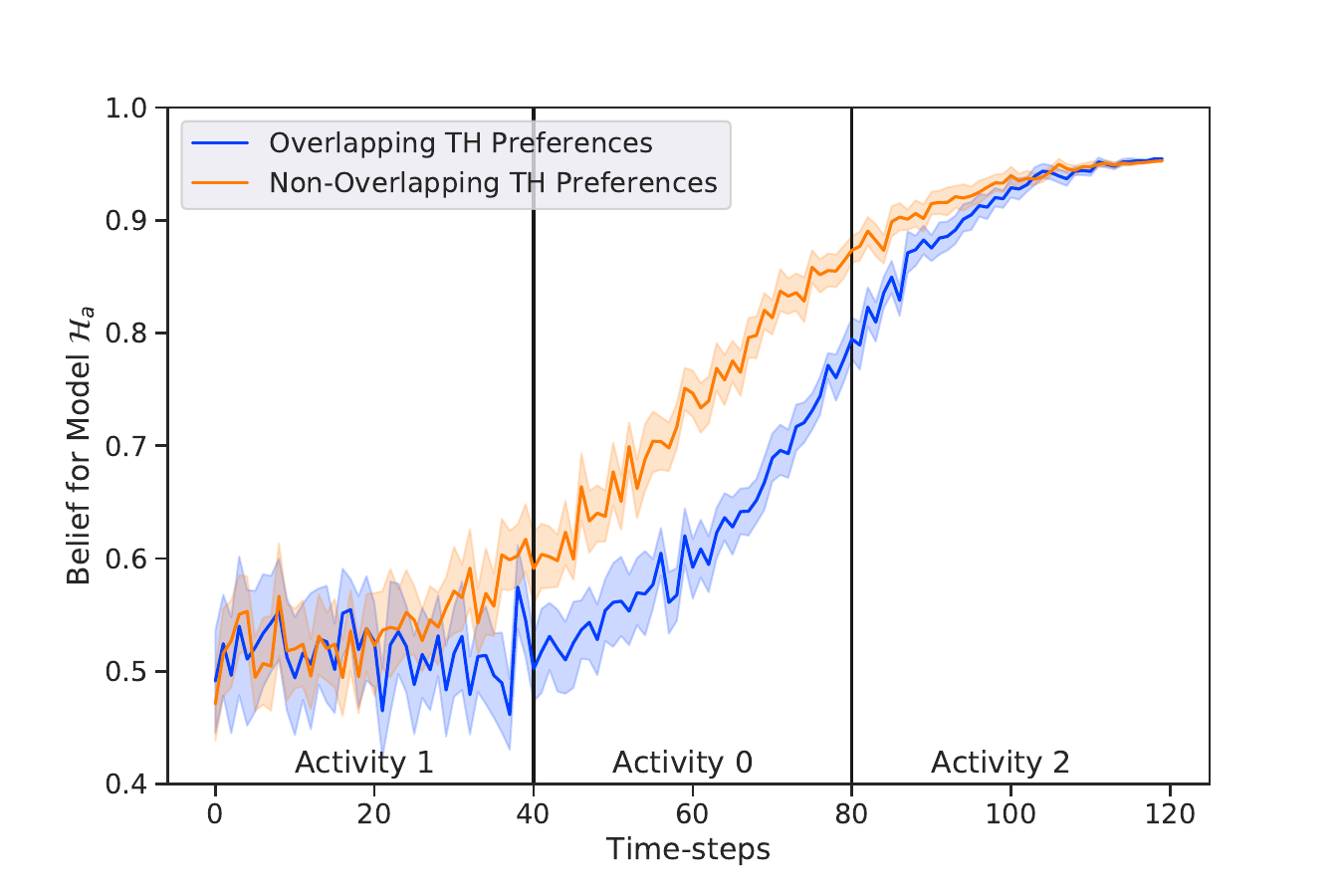}}
    \caption{Model $\mathcal{H}_a$ belief update for overlapping ($d_{0.5}$) and non-overlapping ($d_{0.25}$) TH preferences (with Model $\mathcal{H}_b$). In plot (a), POSHS learns the activity (12-dimension(12d)) specific TH distributions of the occupant while in plot (b), it learns the episode (4-dimension (4d)) specific TH distribution of the occupant.}
   \label{Fig_Belief}
\end{figure}


\begin{figure}[!t]
    \centering
    \subfloat[]{
    \hspace*{-0.2cm}\includegraphics[width=0.55\columnwidth]{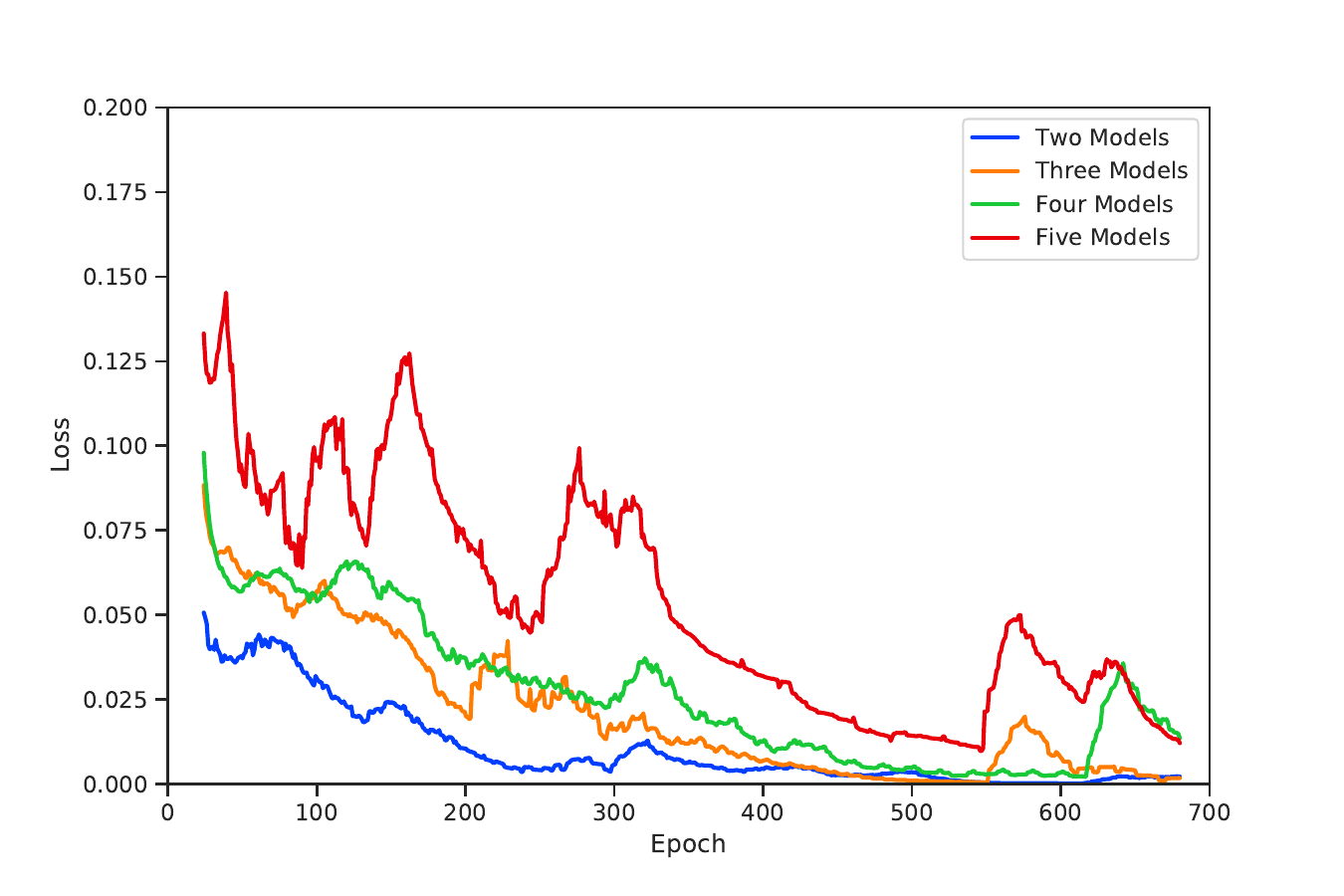}}
    \subfloat[]{
    \hspace*{-0.5cm}\includegraphics[width=0.485\columnwidth]{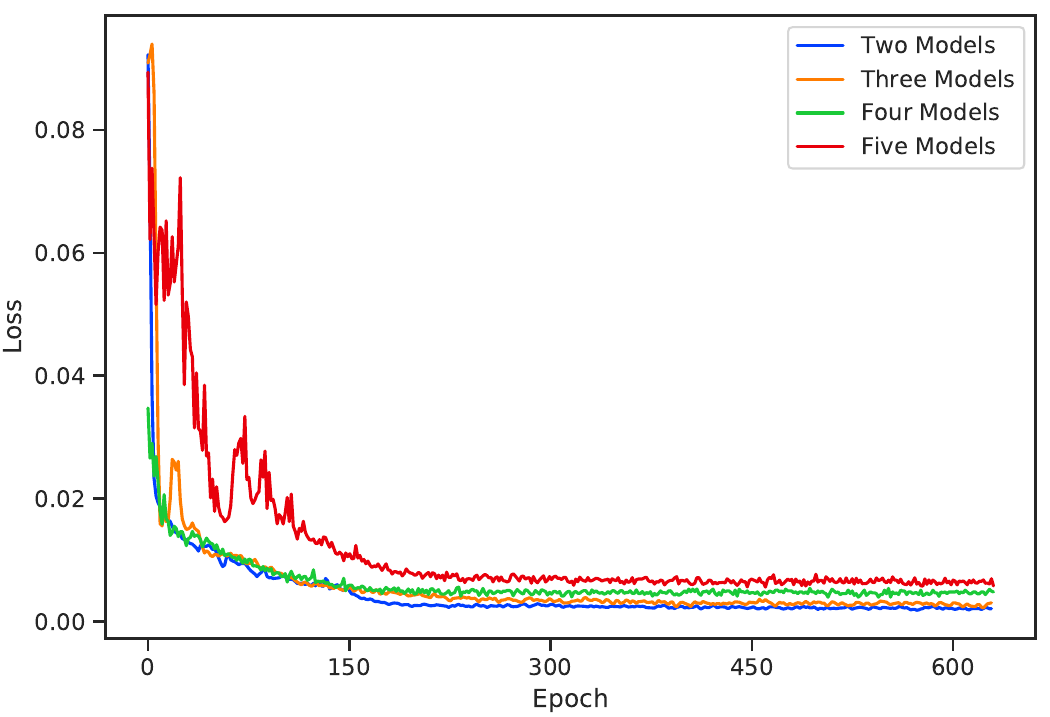}
    }
    \caption{Mean training loss of (a) LSTM and (b) Transformer Attention layer with a combination of two, three, four, and five human models in the environment.}
   \label{Fig_Loss_2345H}
\end{figure}


\begin{figure*}[!t]
    \centering
    \subfloat[]{\includegraphics[width=0.5\columnwidth]{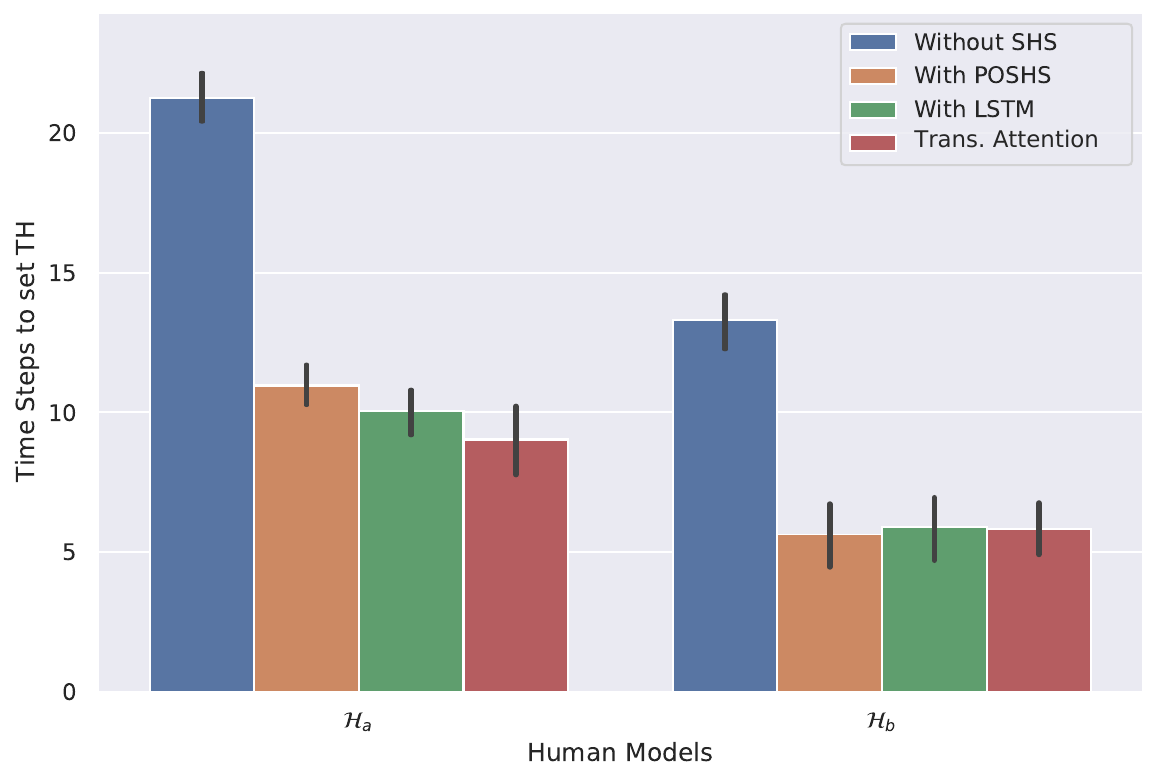}}
    \subfloat[]{\includegraphics[width=0.5\columnwidth]{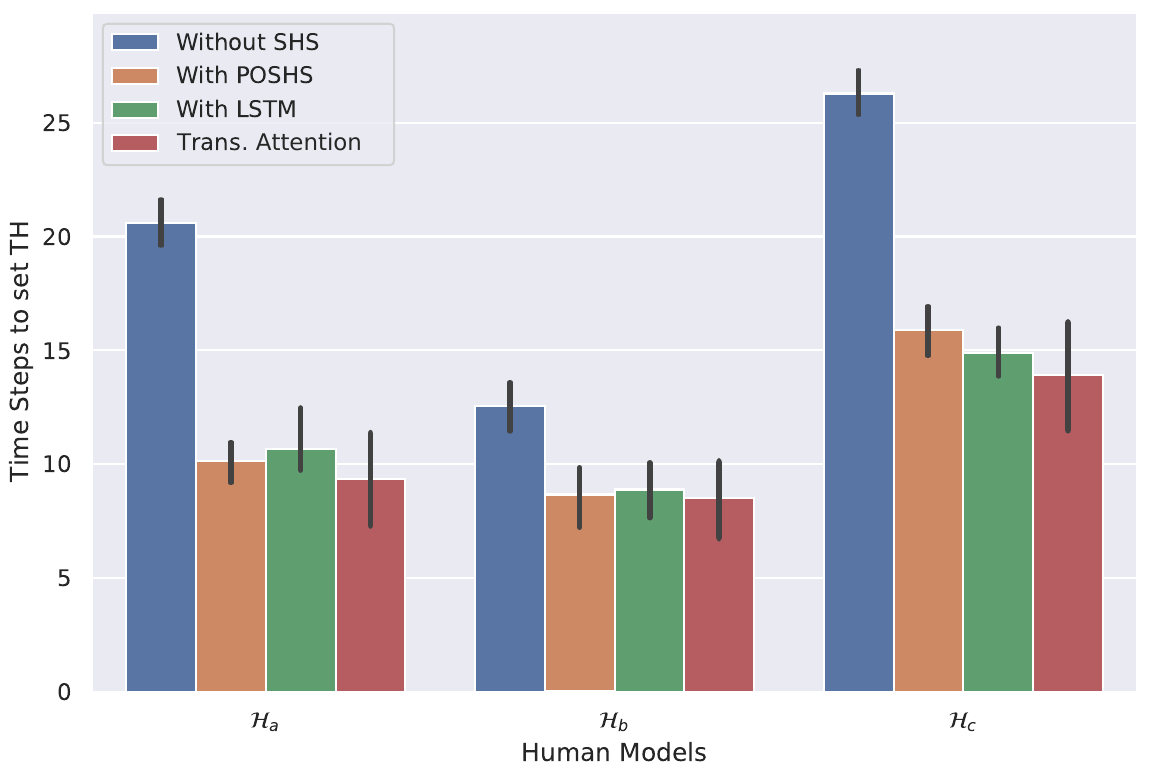}}
    \par
    \subfloat[]{\includegraphics[width=0.5\columnwidth]{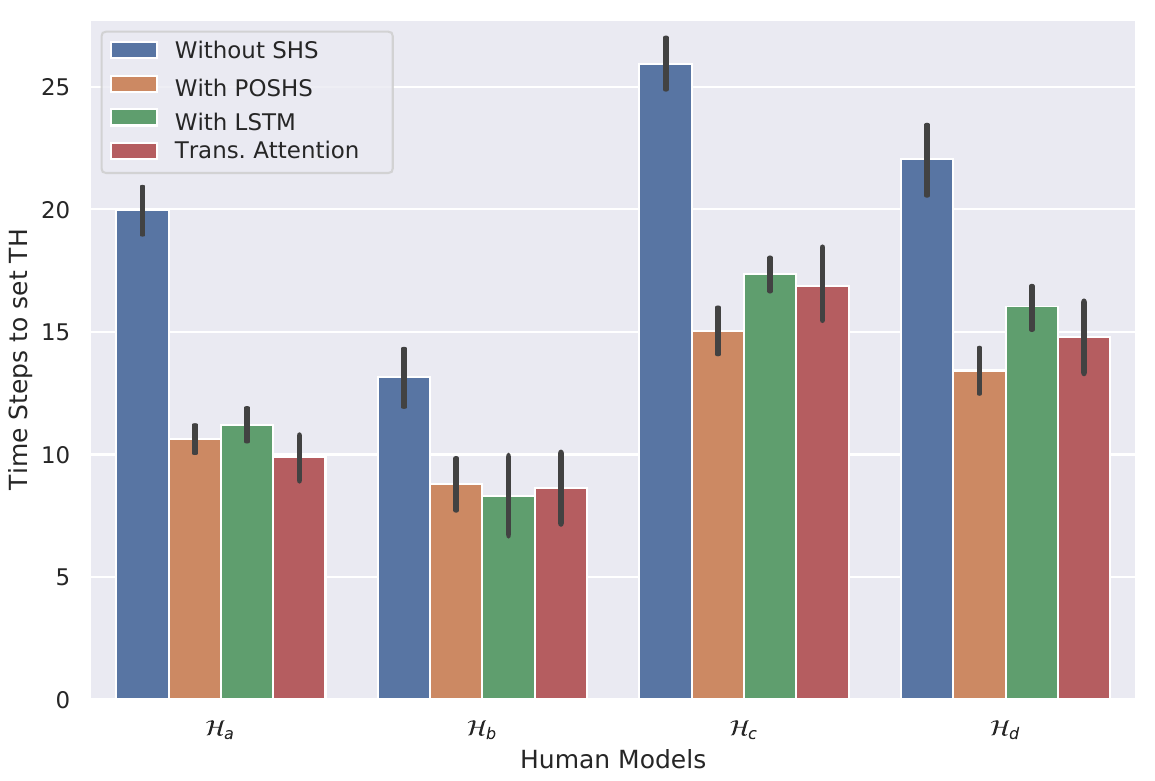}}
    \subfloat[]{\includegraphics[width=0.5\columnwidth]{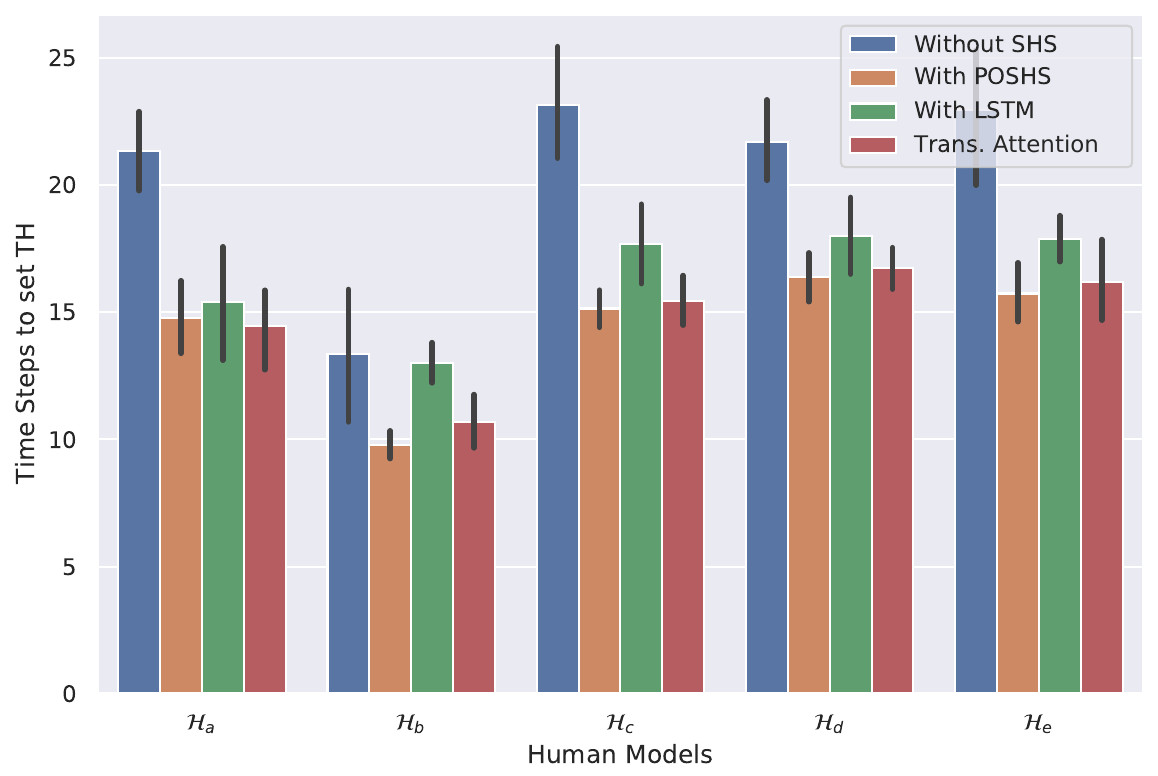}}
    \caption{Mean time-steps taken to set TH without SHS, with POSHS, with LSTM, and with Transformer Attention for a) two, b) three, c) four, and d) five human models in the environment is shown here (lower is better).}
    \label{Fig_TimeSteps_All}
\end{figure*}

To further evaluate the POSHS performance, we repeat the experiment using a wider PMV range of $d_{0.50}$, leading the human models to have more overlapping TH preference distributions. Figure \ref{Fig_Belief}(a) shows the average belief of the POSHS over the occupant during the test episodes. As we can see, when the human model TH preference distributions overlap, it takes more time-steps for the POSHS to properly identify the model's distribution. To further determine if maintaining the per-activity TH preference distributions (12 dimensional: $\mu$ and $\sigma$ for T and H given each activity) is an important factor in the POSHS's performances, we repeat the same experiment once more, but using a TH preference distribution that does not encode activity-specific preferences (4 dimensional: $\mu$ and $\sigma$ for T and H for each episode). The results are shown in Figure \ref{Fig_Belief}(b). As we can see, compared to Figure \ref{Fig_Belief}(a), the two curves are now relatively slow at identifying the current occupant, even when their preference distributions do not overlap as much. For a 12d representation, we re-run preliminary experiments and find $\tau_{JSD} = 0.20$ which empirically showed the best results.

Finally, we repeat the same experiment, using 3, 4, and 5 different human models with a PMV range of $d_{0.25}$ (and the 12d TH preference representation). To do so, we increase the indices of $\mathcal{H}_a$ by small fixed amounts to get the metabolism indices for the new human models $\mathcal{H}_c$, $\mathcal{H}_d$ and $\mathcal{H}_E$. The final values are $[1.15, 1.22, 1.35]$, $[1.15, 1.25, 1.4]$, and $[1.05, 1.3, 1.45]$, respectively. The final accuracies as well as its test F1 scores are reported in Table \ref{Tab_Accuracy}. We observe that with only two human models, the POSHS is able to perform identification with high accuracy based on the thermal preferences. We also observe that with the integration of more human models, the accuracy drops to due to the increased overlap between TH preferences. Nonetheless, strong results (68\%) above chance level (20\%) are still obtained for higher number of human models, for instance 5.

\begin{table*}[!t]
\centering
\caption{POSHS, LSTM and Transformer Attention mean Accuracy (in \%) and F1 Score for accurately approximating the occupant distribution up to 5 human models.(Higher values are better)}
\begin{tabular}{ |c|c|c|c|c|c|c|c|} 
 \hline
  &  \multirow{2}{*}{\textbf{ Model}} & \multicolumn{2}{c|}{\textbf{ POSHS}} &  \multicolumn{2}{c|}{\textbf{ LSTM}} & 
  \multicolumn{2}{c|}{\textbf{ Trans. Att}} \\ 
  & & \textbf{Accuracy} & \textbf{ F1 Score} & \textbf{Accuracy} & \textbf{ F1 Score} &
  \textbf{Accuracy} & \textbf{ F1 Score} \\ \hline
  
 \hline\hline
    \multirow{2}{*}{\textbf{ 2 Models}}
    & Model $\mathcal{H}_a$ & 0.96 & 0.98 & 0.96 & 0.94 & 0.95 & 0.93\\
    & Model $\mathcal{H}_b$ & 1.00 & 0.98 & 0.93 & 0.94 & 0.94 & 0.96\\ \hline
    & \textbf{Mean} & 0.98 & 0.96 & 0.94 & 0.94 & 0.95 & 0.95\\

    \hline \hline
    \multirow{3}{*}{\textbf{ 3 Models}}
    & Model $\mathcal{H}_a$ & 0.92 &  0.89 & 0.94 & 0.93 & 0.93 & 0.88\\
    & Model $\mathcal{H}_b$ & 0.96 &  0.94 & 0.83 & 0.84 & 0.91 & 0.91\\
    & Model $\mathcal{H}_c$ & 0.92 &  0.88 & 0.94 & 0.88 & 0.88 & 0.83\\
 \hline
    & \textbf{Mean} & 0.90  & 0.90 & 0.88 & 0.88 & 0.91 & 0.87\\

    \hline \hline
    \multirow{4}{*}{\textbf{ 4 Models}}
    & Model $\mathcal{H}_a$ & 0.60 & 0.67 & 0.60 & 0.67 & 0.63 & 0.62\\
    & Model $\mathcal{H}_b$ & 0.83 & 0.83 & 0.64 & 0.67 & 0.81 & 0.78\\
    & Model $\mathcal{H}_c$ & 0.90 & 0.78 & 0.89 & 0.76 & 0.81 & 0.77\\
    & Model $\mathcal{H}_d$ & 0.77 & 0.77 & 0.75 & 0.72 & 0.77 & 0.66\\
 \hline
    & \textbf{Mean} & 0.76  & 0.76 & 0.70 & 0.71 & 0.76 & 0.71\\
    
    \hline \hline    
    \multirow{5}{*}{\textbf{ 5 Models}}
    & Model $\mathcal{H}_a$ & 0.67 & 0.63 & 0.78 & 0.74 & 0.68 & 0.69 \\
    & Model $\mathcal{H}_b$ & 0.60 & 0.60 & 0.50 & 0.55 & 0.67 & 0.60 \\
    & Model $\mathcal{H}_c$ & 0.88 & 0.78 & 0.75 & 0.67 & 0.81 & 0.71 \\
    & Model $\mathcal{H}_d$ & 0.70 & 0.70 & 0.56 & 0.50 & 0.59 & 0.66 \\
    & Model $\mathcal{H}_e$ & 0.62 & 0.70 & 0.62 & 0.70 & 0.68 & 0.67 \\
 \hline
    & \textbf{Mean} & 0.67  & 0.68 & 0.62 & 0.63 & 0.69 & 0.67\\
    \hline 
    \end{tabular}
\label{Tab_Accuracy}
\end{table*}

\begin{table}[!t]
\centering
\caption{Mean Reward values for up to 5 human models for the POSHS, LSTM, and Transformer Attention baseline in comparison to when the SHS is not integrated (higher values are better).}
\begin{tabular}{ |c|c|c|c|c| } 
    \hline
    {\textbf{\footnotesize Model}} & {\textbf{\footnotesize Without SHS}} & {\textbf{\footnotesize POSHS}} & {\textbf{\footnotesize LSTM}} & {\textbf{\footnotesize Trans. Att.}}\\
    \hline\hline
    \multirow{1}{*}{\textbf{{\footnotesize 2 Models}}} 
    & 284$\pm$1.17 & 291$\pm1.14$ & 290$\pm1.62$ & 293$\pm$2.73\\
    \hline
    \multirow{1}{*}{\textbf{{\footnotesize 3 Models}}} 
    & 283$\pm$1.80 & 291$\pm1.81$ & 286$\pm2.78$ & 291$\pm$3.01\\
    \hline
    \multirow{1}{*}{\textbf{{\footnotesize 4 Models}}} 
     & 283$\pm$1.92 & 287$\pm3.49$ & 284$\pm3.89$ & 289$\pm$2.71\\
    \hline
    \multirow{1}{*}{\textbf{{\footnotesize 5 Models}}} 
     & 283$\pm$2.35 & 286$\pm4.30$ & 283$\pm5.70$ & 283$\pm$6.41\\
    \hline
\end{tabular}
\label{Tab_Reward}
\end{table}

\subsection{Experiment B}\label{Exp_B}

Here, we evaluate how both approaches, the POSHS and the LSTM model, perform in terms of rewards and time steps required for the human occupant to set TH as we increase the number of humans models in the home. Note that both POSHS and LSTM must form an internal representation of the underlying hidden state. POSHS does it by explicitly estimating the current occupant's TH preference, while the LSTM builds an internal representation based on the sequence of observed actions (TH and activities). 

Similar to Experiment A, we train the LSTM with the trained human occupant for 175 episodes with a learning rate of $0.0013$. During the training, a human model is chosen randomly at the beginning of the episode. We repeat the experiment  for up to 5 human models. For testing purposes, we test the LSTM model for 50 episodes to evaluate the final performance. Therefore, both POSHS and LSTM were tested under the same training and testing conditions.

The performance in terms of the occupant's rewards for both SHS models are shown in Table \ref{Tab_Reward}. There is a slight increase in the observed mean reward for the occupants with the SHS in comparison to mean reward without the SHS, suggesting that the human occupants spend less time changing the TH in the presence of the SHS model.

We also compare the user recognition ability of POSHS and LSTM model of SHS. POSHS recognizes the current user by the selecting the TH distribution that has the highest belief. It also checks if the current user's $JSD < \tau_{JSD}$ for the distribution that has the highest belief. Similarly, the LSTM based model recognizes the user by comparing the embedding of its hidden state with the embeddings of previously observed users using KST as described in section IV. Table \ref{Tab_Accuracy} shows the identification accuracy by the POSHS and LSTM model. We observe that for 2-3 human models in the environment, the baseline LSTM is only slightly behind POSHS. However, with more human models, the difference increases.

We then measure the actual number of time-steps required by the occupants to set their preferences with the SHS models compared to the time-steps required by the occupants without the SHS. This is shown in Figure \ref{Fig_TimeSteps_All} where we present the results for 2, 3, 4, and 5 human occupants in the SHS. It is observed that the number of extra time steps taken by the occupants are close for POSHS and LSTM for up to 3 occupants. However, when compared to Figure \ref{Fig_TimeSteps_All}(a) and (b) in Figure \ref{Fig_TimeSteps_All}(c) and (d), the difference between the time-steps of LSTM and POSHS increases with more human occupants. Furthermore, referring back to Table \ref{Tab_Reward}, we observe that POSHS slightly outperforms the LSTM model as the amount of reward received by the occupant with POSHS is higher, and the difference increases as the number of human models increases. 

The LSTM training loss curve is presented in Figure \ref{Fig_Loss_2345H}, showing that the LSTM struggles to learn the $Q$-values accurately as the number of occupants increases. This is particularly highlighted with 4 and 5 human models, where the LSTM seems to suffer from considerable forgetting \cite{Forget} (depicted by the peaks in the training loss curve) due to the changing weights of the network required to learn the sequences of different TH preferences.

\subsection{Experiment C} Here, we modify our LSTM baseline by replacing the LSTM layer with a Transformer Attention layer. Similar to the LSTM model, this new baseline also has to learn the internal representation of the human using its activity and thermal preferences. We train this model for 175 episodes with a learning rate of 0.0013. We perform the experiment with up to 5 human models. In order to test the model's performance, we evaluate the trained model with trained human models for 50 episodes. 
In order to identify the current occupant, we use the encoding obtained from the fully connected layer of our Transformer model for each occupant. KST as described in section IV is then used to identify the similarity between the current encoding and previously observed encodings. Table \ref{Tab_Accuracy} shows the identification accuracy of POSHS, LSTM, and Transformer Attention.

Figure \ref{Fig_TimeSteps_All} shows the comparison of the time-steps required by the occupant to set TH in the presence of POSHS, LSTM, and Transformer Attention. We observe that for up to 3 human models, the Transformer Attention performs slightly better than the LSTM baseline and comparable to the POSHS model. This is due to the ability of Attention \cite{transformer} in processing non-sequential states, which means that the order of activity execution by the human model does not matter unlike for the LSTM model. With 4 or 5 human models, the Transformer Attention baseline outperforms the LSTM and achieves competitive results with respect to the POSHS model. Nonetheless, it should be mentioned that our POSHS model, like the LSTM, only carries out a simple update at each time step, while the Transformer Attention needs to re-read the entire sequence at each time step, thus requiring quadratic computations with respect to the sequence length. Moreover, our POSHS model has 10,125 parameters (for 5 human models), while the LSTM network contains 146,256 parameters (weights), and the Transformer Attention network consists of 193,832 parameters.

Figure \ref{Fig_Loss_2345H}(b) shows the loss curves for the Transformer Attention model, which is slightly better than the LSTM curves where occasional spikes were observed as the number of humans models increased and the order of activities were changed. For 2-3 human models, the loss decreases sharply unlike for 4-5 human models where the converging loss is slightly greater. 


\section{Conclusions and Future Work}
In this paper, we investigate the scenario where limited information about the occupants is available to the smart home, resulting is often sub-optimal actions by the smart home in a given state. We model a smart home to learn to recognize the occupant's thermal preference distributions using Bayesian modelling, which helps maintain a belief over the hidden user state that the smart home uses to improve its policy. As a baseline for comparison, we design an approach in which the temporal relations between the TH sequences are learned to approximate the hidden state. We integrate both smart home models with up to 5 human models and evaluate them based on their ability to identify the underlying state (or occupant's TH preferences), as well as the overall human performance in terms of human rewards and time steps spent changing the TH with the SHS. Finally, we compare the POSHS with the baseline model that aims to learn its own representation of underlying states of the partially observable environment. Our simulations show good performance with POSHS when the number of human occupants are low without an increase in time required to set the thermal preferences by the occupant. Similar performance was observed with the baseline. With more human models integrated in the environment, the time-step difference between the baseline and POSHS increases where the occupants now take more time to set the TH with the baseline compared to the POSHS model. In the end, With our simulated experiments, we demonstrate that in an environment where the user information is not fully observable it is possible to approximate the user's hidden state with multiple occupants without having an impact on the user's behavior even with sub-optimal approximations.

For future work, we may include additional occupant-related parameters in our algorithm, which can help in better approximation of the occupant and thus improve performance. These parameters may include heart rate, skin temperature, skin conductance, surrounding radiation, and others. Some of the most common bio-signals like Photoplethysmogram (PPG), Electrocardiogram (ECG), Electrodermal Activity (EDA), and Electroencephalogram (EEG) can be obtained by smart wearables that have currently become very popular. This will not only improve our algorithm's accuracy, but will also provide more data to work with, which can allow for our algorithm to scale to more than 5 humans. Having multiple occupants at the same time also increases the complexity of the problem. However, by recognizing the users, the methods proposed in previous work such as \cite{Suman} can also be readily employed.

\small
\bibliographystyle{abbrv}
\bibliography{references}
\end{document}